%% file: main.tex
\documentclass[10pt,twocolumn,letterpaper]{article}

\usepackage[pagenumbers]{cvpr} 

\input{math_commands}

\definecolor{cvprblue}{rgb}{0.21,0.49,0.74}
\definecolor{myred}{HTML}{F54254}
\definecolor{myorange}{HTML}{FFB135}
\definecolor{mygreen}{HTML}{10BD35}
\definecolor{lygreen}{HTML}{4CC764}
\definecolor{myblue}{HTML}{598BE7}
\definecolor{mypurple}{HTML}{9A1C6B}
\definecolor{plgray}{HTML}{999999}
\definecolor{motion}{rgb}{0.95,0.65,0.25}
\definecolor{hint}{rgb}{0.72,0.72,0.72}
\usepackage[pagebackref,breaklinks,colorlinks,allcolors=cvprblue]{hyperref}
\usepackage[most]{tcolorbox}
\usepackage{makecell}
\usepackage{multirow}
\usepackage{algorithm,algorithmic}
\usepackage{soul}
\usepackage{lipsum}
\usepackage{tikz}

\makeatletter
\g@addto@macro\normalsize{%
  \setlength\abovedisplayskip{6pt}
  \setlength\belowdisplayskip{6pt}
  \setlength\abovedisplayshortskip{4pt}
  \setlength\belowdisplayshortskip{4pt}
}
\makeatother

\def\paperID{*****} 
\def\confName{CVPR}
\def\confYear{2026}

\title{ProjFlow: Projection Sampling with Flow Matching for \\ Zero‑Shot Exact Spatial Motion Control}

\author{
Akihisa Watanabe\textsuperscript{1*}\quad
Qing Yu\textsuperscript{2}\quad
Edgar Simo-Serra\textsuperscript{1}\quad
Kent Fujiwara\textsuperscript{2}\\[0.5em]
\textsuperscript{1}Waseda University\qquad
\textsuperscript{2}LY Corporation
}

\begin{document}
\twocolumn[{
\renewcommand\twocolumn[1][]{#1}
\maketitle
\begin{center}
    \centering
    \captionsetup{type=figure}
    \includegraphics[width=\textwidth]{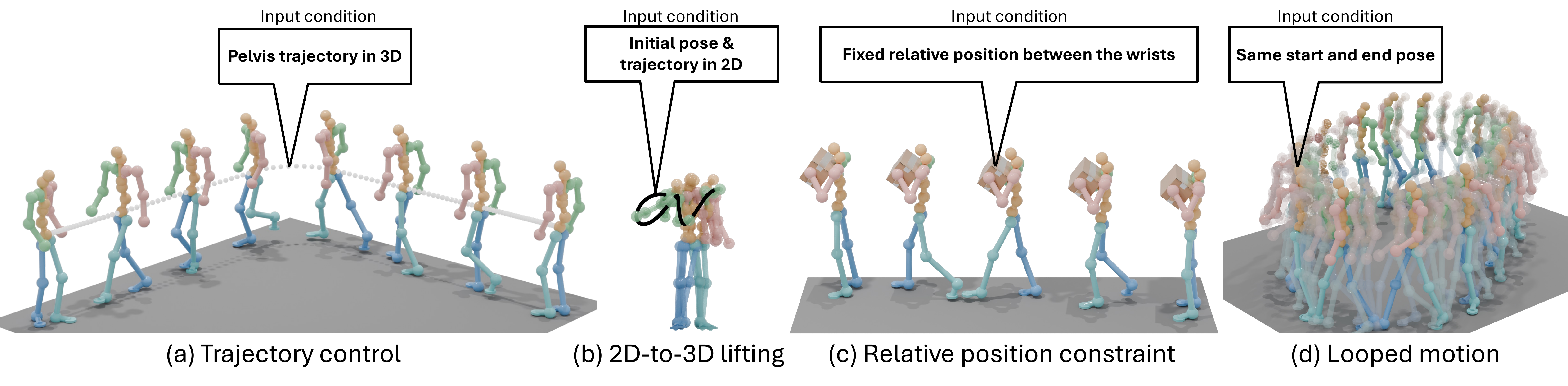}
    \vspace{-2.0em}
    \caption{\textbf{ProjFlow} provides a unified, zero-shot framework for exact spatial motion control. The method handles diverse applications by formulating them as linear inverse problems. Examples of applications include (a) precisely following a specified joint's trajectory, (b) lifting 2D keypose and 2D trajectory inputs to a full 3D motion, (c) maintaining a fixed relative position between joints, and (d) generating seamlessly looped motion by matching start and end poses.}
    \label{fig:teaser}
\end{center}
}]
\begingroup
\renewcommand{\thefootnote}{\fnsymbol{footnote}}
\footnotetext{*Work done during an internship at LY Corporation.}
\endgroup
\setcounter{footnote}{0}

\input{sec/0_abstract}    

\input{sec/1_intro}
\input{sec/2_related_work}
\input{sec/3_preliminaries}
\input{sec/4_method}
\input{sec/5_experiment}
\input{sec/6_limitations}
\input{sec/7_conclusion}

{
    \small
    \bibliographystyle{ieeenat_fullname}
    \bibliography{main}
}

\appendix
\input{sec/suppl}

\end{document}

%% file: math_commands.tex
\usepackage{amsmath,amsfonts,bm}

\def\eqref#1{equation~\ref{#1}}

\def\1{\bm{1}}

\def\vd{{\bm{d}}}
\def\ve{{\bm{e}}}

\def\vx{{\bm{x}}}
\def\vy{{\bm{y}}}

\def\vepsilon{{\bm{\epsilon}}}

\DeclareMathAlphabet{\mathsfit}{\encodingdefault}{\sfdefault}{m}{sl}
\SetMathAlphabet{\mathsfit}{bold}{\encodingdefault}{\sfdefault}{bx}{n}

\newcommand{\E}{\mathbb{E}}

\DeclareMathOperator*{\argmin}{arg\,min}

\usepackage{capt-of}
\usepackage{diagbox}

\definecolor{C0}{rgb}{0.121569, 0.466667, 0.705882}
\definecolor{C1}{rgb}{1.000000, 0.498039, 0.054902}
\definecolor{C2}{rgb}{0.172549, 0.627451, 0.172549}
\definecolor{C3}{rgb}{0.839216, 0.152941, 0.156863}
\definecolor{C4}{rgb}{0.580392, 0.403922, 0.741176}
\definecolor{C5}{rgb}{0.549020, 0.337255, 0.294118}
\definecolor{C6}{rgb}{0.890196, 0.466667, 0.760784}
\definecolor{C7}{rgb}{0.498039, 0.498039, 0.498039}
\definecolor{C8}{rgb}{0.737255, 0.741176, 0.133333}
\definecolor{C9}{rgb}{0.090196, 0.745098, 0.811765}
\definecolor{trolleygrey}{rgb}{0.5, 0.5, 0.5}

\definecolor{BrickRed}{rgb}{0.6,0,0}
\definecolor{RoyalBlue}{rgb}{0,0,0.8}
\definecolor{Tdgreen}{rgb}{0,0.4,0.7}
\definecolor{pinegreen}{rgb}{0.0, 0.47, 0.44}
\definecolor{cornellred}{rgb}{0.7, 0.11, 0.11}
\definecolor{cadmiumgreen}{rgb}{0.0, 0.42, 0.24}
\definecolor{spirodiscoball}{rgb}{0.06, 0.75, 0.99}
\definecolor{mylightblue}{rgb}{0.85, 0.90, 0.94}
\definecolor{maroon}{cmyk}{0,0.87,0.68,0.32}

\usepackage{pifont}
\definecolor{cfg}{rgb}{0.906, 0.435, 0.318}
\definecolor{cfgpp}{rgb}{0.165, 0.616, 0.561}
\definecolor{cfgnull}{rgb}{0.208, 0.565, 0.953}



%% file: sec/0_abstract.tex
\begin{abstract}
Generating human motion with precise spatial control is a challenging problem. Existing approaches often require task-specific training or slow optimization, and enforcing hard constraints frequently disrupts motion naturalness. Building on the observation that many animation tasks can be formulated as a linear inverse problem, we introduce \textbf{ProjFlow}, a training-free sampler that achieves zero-shot, exact satisfaction of linear spatial constraints while preserving motion realism. Our key advance is a novel kinematics-aware metric that encodes skeletal topology. This metric allows the sampler to enforce hard constraints by distributing corrections coherently across the entire skeleton, avoiding the unnatural artifacts of naive projection. Furthermore, for sparse inputs, such as filling in long gaps between a few keyframes, we introduce a time-varying formulation using pseudo-observations that fade during sampling. Extensive experiments on representative applications, motion inpainting, and 2D-to-3D lifting, demonstrate that ProjFlow achieves exact constraint satisfaction and matches or improves realism over zero-shot baselines, while remaining competitive with training-based controllers. 
\end{abstract}

%% file: sec/1_intro.tex
\vspace{-1em}
\section{Introduction}
\label{sec:intro}
An open challenge in character animation is spatial motion control, which involves generating realistic full-body motion that conforms to user-defined spatial cues. These cues can include trajectories, target poses, or specific joint locations. Solving this task would allow 3D animators to work with precise and interactive control, immediately obtaining desired motions that remain natural and diverse~\citep{studer2024factorized, agrawal2024skelbetweener}.

Users typically specify constraints for only a subset of the body, such as the trajectory of a single hand or foot. This makes the spatial motion control problem ill-posed, with many motions satisfying these sparse constraints. An intuitive approach to resolve this ambiguity is to favor motions with high likelihood under a pretrained motion prior, selecting the most natural result from all valid options.

Building on this idea, dominant approaches steer pretrained diffusion models to satisfy user-defined spatial constraints. However, existing methods suffer from significant limitations. They often require task-specific training for conditioning branches~\cite{xie2024omnicontrol,meng2025acmdm,dai2024motionlcm,pinyoanuntapong2024controlmm}, or they rely on slow, inference-time optimization~\cite{karunratanakul2024dno,ron2025hoidini,pi2025coda,pinyoanuntapong2024controlmm}, which reduces interactivity and can get stuck in local minima. Fundamentally, these approaches treat constraints as soft objectives rather than hard rules. As a result, exact satisfaction is not guaranteed, and residual violations persist. What is missing is a sampler that can (i) enforce hard equality constraints exactly, (ii) operate zero-shot without task-specific retraining, and (iii) require no inner-loop optimization at inference time, all while preserving the pretrained motion prior.

In this paper, we present \textbf{ProjFlow}, \textbf{Proj}ection Sampling with \textbf{Flow} Matching for zero-shot exact spatial motion control. We begin with the observation that a wide range of motion control and editing tasks can be formulated as linear inverse problems. These tasks include trajectory following, keyframing, camera or root path control, and partial-body editing. ProjFlow addresses these problems by projecting the predicted clean motion at every denoising step onto the set of motions that satisfy the given constraints. This projection introduces the smallest necessary adjustment, measured under a newly designed \emph{kinematics-aware metric} that reflects skeletal topology. Rather than measuring the distance in Euclidean space, this metric ensures that updates propagate coherently along the kinematic tree, avoiding unnatural and isolated joint movements. Hard constraints are satisfied exactly, while uncertain or partial measurements are weighted according to their confidence. The projected update is then combined with a flow-matching recomposition step, preserving the pretrained motion prior without any task-specific retraining or inner-loop optimization.

We evaluate the versatility of the ProjFlow framework through two representative applications in spatial motion control. The first application is motion inpainting, where segments of a motion sequence are entirely missing. This task requires the model to infer plausible intermediate frames from sparse temporal observations. Instead of treating the unobserved frames as blanks, ProjFlow introduces pseudo-observations around known frames and gradually adjusts their influence during sampling, enabling coherent zero-shot completion even across long temporal gaps.

The second application is 2D-to-3D motion reconstruction, where the input consists of 2D keypoints and their trajectories over time. The goal is to recover the underlying 3D motion that projects onto the observed 2D data. ProjFlow enforces linear measurement constraints derived from the camera model as hard equalities at each step. This yields accurate 3D reconstructions with zero reprojection error and natural motion. Our experiments on these applications show ProjFlow matches the accuracy of training-based methods without any retraining or inner-loop optimization. These results demonstrate the versatility of our framework, which can also be applied to the other tasks illustrated in Fig.~\ref{fig:teaser}.

In summary, our contributions are as follows:
\begin{itemize}
\setlength\itemsep{0pt}
\item \textbf{Unified linear inverse formulation and projection sampler as its solver.} We cast motion control and editing as linear inverse problems and propose a projection-based flow-matching sampler that enforces constraints exactly without retraining or inner-loop optimization.

\setlength\itemsep{0pt}
\item \textbf{Kinematics-aware projection geometry.} We introduce a metric that encodes skeletal structure, providing a principled geometry that distributes corrections coherently and improves realism and stability.

\setlength\itemsep{0pt}
\item \textbf{Empirical parity on inpainting and 2D-to-3D with exact constraints.} Through experiments on motion inpainting and 2D-to-3D reconstruction, we show that ProjFlow matches the performance of training-based models while satisfying the specified constraints exactly up to numerical precision, all in a zero-shot, no inner loop setting.
\end{itemize}

%% file: sec/2_related_work.tex
\vspace{-0.3em}
\section{Related Work}
\vspace{-0.2em}
\subsection{Human Motion Generation}
\vspace{-0.2em}
Recent advances in image generation indicate a transition from denoising diffusion probabilistic models and score-based SDEs to flow matching models that learn velocity fields using rectified-flow objectives, scaling well with Transformer architectures~\citep{ddpm_ho2020,score_song2021,flow_lipman2023,flow_liu2023,sd3_esser2024,flow_lipman2024}. Progress in text-conditioned human motion generation has followed the same arc. Early state-of-the-art systems were diffusion-based~\citep{tevet2023human,zhang2024motiondiffuse,dabral2023mofusion,zhang2023remodiffuse}, while more recent work adopts flow-matching formulations~\citep{hu2023mfm,cuba2025flowmotion}.

Alongside advances in generative methodology, motion representation has also evolved. HumanML3D~\citep{guo2022humanml3d} popularized a kinematic, relative, and partly redundant feature representation still adopted by many controllers~\citep{guo2022humanml3d,xie2024omnicontrol,karunratanakul2023gmd,dai2024motionlcm}. Evidence now shows that generating absolute joint coordinates in world space with a rectified-flow objective is effective and beneficial for controllability and scalability~\citep{meng2025acmdm,meng2025rethinking}. These trends motivate our choice of a flow-matching sampler operating directly in world coordinates.

\vspace{-0.3em}
\subsection{Spatially Controlled Motion Generation}
\vspace{-0.3em}
While text prompts are effective for controlling high-level motion semantics, many practical applications require more precise spatial control. Synthesizing motion from a wider range of external control signals, often in combination with text prompts, has been widely explored. Examples include authoring from storyboard sketches \citep{zhong2025sketch2anim} and multi-track timeline authoring \citep{petrovich2024multitrack}. Other research streams focus on multi-objective control for characters and robots \citep{serifi2024robotmdm,alegre2025amor}, music-conditioned choreography \citep{lee2019dancing,li2021aichoreographer,tseng2022edge,li2022bailando,li2023bailandopp,li2021danceformer}, or generating motions involving inter-human~\cite{tanaka2023interaction,liang2024intergen,fan2024freemotion,ota2025pino} and human-object interactions \citep{cha2024text2hoi,diller2024cghoi,kulkarni2024nifty,li2023controllableHOI}. Control signals can also include sparse tracking inputs \citep{du2023avatars}, scene affordances \citep{huang2023diffusion3d,wang2024moveasyousay}, programmable objectives \citep{liu2024progmogen}, style specifications \citep{zhong2024smoodi}, or goal-directed targets \citep{diomataris2024wandr}.

A key question is how to effectively integrate these spatial signals into text-to-motion generators to enforce precise accuracy. Prior work has taken several routes to tackle this. One approach involves fine-tuning diffusion priors with end-effector supervision \citep{shafir2024priormdm} or training models for in-betweening from dense or sparse keyframes \citep{cohan2024condmdi}. Another line of work applies guidance during sampling, steering the generation towards root or waypoint trajectories \citep{karunratanakul2023gmd,rempe2023tracepace}. More recently, joint-wise conditioning has been achieved using ControlNet-style branches or latent controllers \citep{xie2024omnicontrol,dai2024motionlcm,zhang2023controlnet}. Others perform inference-time optimization of the initial noise or logits to minimize differentiable objectives \citep{karunratanakul2024dno,pinyoanuntapong2024controlmm}, or use factorization and controller mixtures for fine-grained control \citep{wan2024tlcontrol,liang2024omg}. Across these routes, constraints are injected as differentiable penalties or guidance terms rather than enforced as hard feasibility constraints. Consequently, exact feasibility is not guaranteed, and methods often require task‑specific conditioning or iterative inner‑loop optimization during inference.

\vspace{-0.3em}
\subsection{Inverse Problems with Image Generation}
\vspace{-0.2em}
Pre-trained diffusion priors have enabled strong zero-shot solvers for linear inverse problems. Two influential views have emerged. The first is likelihood guidance along the sampling path~\citep{chung2023diffusion,kawar2022denoising}. The second is \emph{projection} that freezes range-space and refines only the null-space (DDNM)~\citep{wang2023zeroshot}, with extensions such as pseudoinverse guidance~\citep{song2023pseudoinverseguided}. To leverage large latent generative models, latent diffusion model-based variants inject data consistency in latent space \citep{rout2023solving,song2024solving,zhang2024improving}. Recently, these ideas have been extended to flow models. FlowChef and PnP-Flow steer rectified-flow fields or plug a learned denoiser into a flow solver \citep{patel2024steering,martin2025pnpflow}, but do not cast inverse solving as closed-form posterior steps on the flow path.

ProjFlow adapts data consistency updates to the flow matching regime, and the framework generalizes prior posterior projection samplers in two key ways. First, it replaces the common Euclidean geometry of image methods with a kinematics-aware metric that distributes corrections coherently along the skeleton, which better supports structured data such as human motion. Second, the framework introduces time-scheduled pseudo-observations that densify guidance in unobserved regions and then fade as sampling proceeds, improving on prior approaches that treat missing regions as simple blanks. Finally, ProjFlow recovers DDNM in the Euclidean noiseless deterministic limit while extending support to structured metrics, noisy measurements, and time-varying operators.

%% file: sec/3_preliminaries.tex
\vspace{-0.6em}
\section{Preliminaries}
\vspace{-0.2em}
\subsection{Motion Representation}
We represent a clean motion sequence of length $N$ with $J$ joints in absolute world coordinates as a tensor
$\vx \in \mathbb{R}^{N\times J\times 3}$.
For brevity, we also use $\vx$ to denote its vectorization $\vx \in \mathbb{R}^d$ with $d=3JN$. Unless stated otherwise, we assume a frame-major order. Each vector element $i\in\{1,\dots,d\}$ corresponds to a unique frame–joint–spatial-channel triple $(n_i, j_i, c_i)$, where $n_i\!\in\!\{1,\dots,N\}$, $j_i\!\in\!\{1,\dots,J\}$, and $c_i\!\in\!\{x,y,z\}$.

\subsection{Flow Matching}
\label{sec:prelim}
The core idea of flow-based generative models~\cite{flow_lipman2023, flow_liu2023, flow_albergo2023} is to learn a time-dependent vector field $v_\theta(\vx, t)$ that transports samples from a simple prior distribution $p_0$ to a complex target data distribution $q$.

Let $\psi_t: \mathbb{R}^d \!\to\! \mathbb{R}^d$ denote the flow map induced by this vector field. The flow map is defined as the unique solution to the Ordinary Differential Equation (ODE)
\begin{equation}
\frac{d\psi_t(\vx_0)}{dt} \;=\; v_\theta\!\big(\psi_t(\vx_0), t\big), 
\qquad 
\psi_0(\vx_0) = \vx_0,
\label{eq:ode}
\end{equation}
where $\vx_0$ is the initial condition.

In this study, we adopt the Rectified Flow formulation~\cite{flow_liu2023, flow_lipman2024}, which defines a straight-line path between a noise sample $\vx_0$ and a data sample $\vx_1$:
\begin{equation}
    \vx_t = (1-t)\,\vx_0 + t\,\vx_1, \qquad t \in [0,1].
    \label{eq:linear_path}
\end{equation}
Along this path, the ideal velocity is constant and equal to $\vx_1 - \vx_0$. The network $v_\theta$ is trained to approximate the conditional expectation of this velocity given $(\vx_t, t)$ by minimizing the conditional flow-matching loss
\begin{equation}
    \mathcal{L}_{\text{FM}}(\theta)
    = \mathbb{E}_{\substack{
        t \sim \mathcal{U}(0,1)\\
        \vx_0 \sim p_0\\
        \vx_1 \sim q
    }}
    \!\left[\,\big\| v_\theta(\vx_t, t) - (\vx_1 - \vx_0) \big\|_2^2 \right], 
    \label{eq:fm_loss}
\end{equation}
where $\vx_t$ is given by~\eqref{eq:linear_path}. Sampling is then performed by drawing $\vx_0 \sim p_0$ and numerically integrating the ODE in~\eqref{eq:ode} from $t=0$ to $t=1$ to obtain $\vx_1 = \psi_1(\vx_0)$.

This formulation provides a continuous and differentiable generative path between the prior and data distributions, which later facilitates direct constraint enforcement in our projection-based framework.

%% file: sec/4_method.tex
\begin{figure*}[t]
    \centering
    \includegraphics[width=\linewidth]{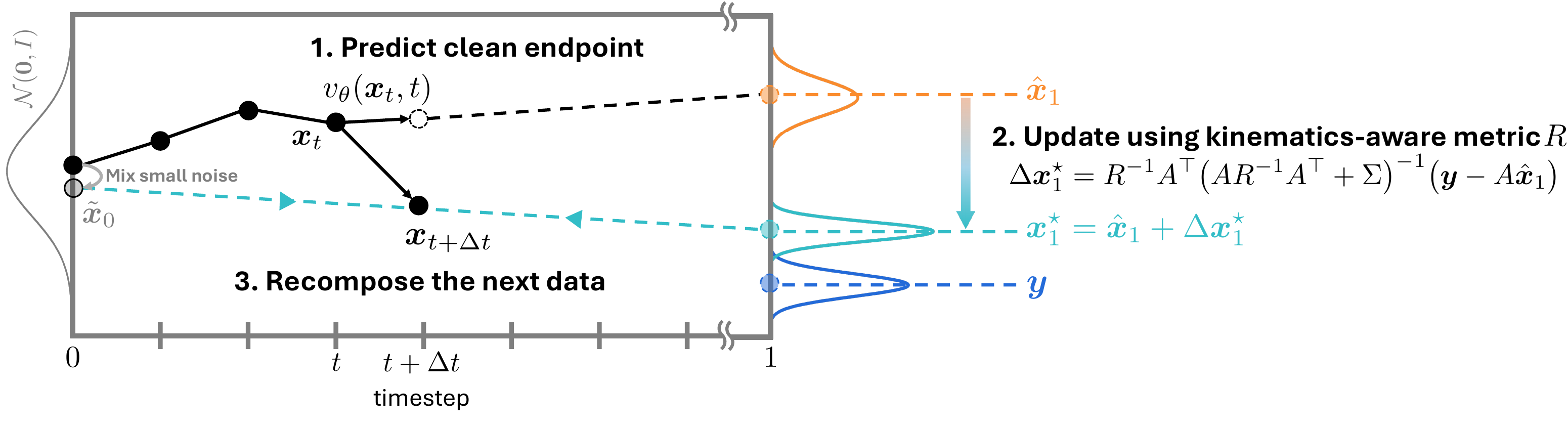}
    \vspace{-2.0em}
    \caption{\textbf{Overview of the Projection Sampling Step.}
    At each timestep $t$: (1) predict the clean endpoint $\hat{\vx}_1$ from $\vx_t$ using the learned velocity $v_\theta(\vx_t,t)$; (2) enforce the linear–Gaussian measurements $\vy=A\vx+\vepsilon$ by computing a correction $\Delta \vx_1^\star$ that projects $\hat{\vx}_1$ to the measurement set under the \emph{kinematics-aware} metric $R$. This metric encodes skeletal topology and spreads updates coherently along the kinematic tree. The measurement covariance $\Sigma$ modulates the pull toward the observations; smaller values yield stronger attraction and recover hard constraints as $\Sigma \to 0$. (3) Finally, stochastically recompose the corrected endpoint to obtain the next state $\vx_{t+\Delta t}$.}
    \label{fig:projflow}
\end{figure*}

\section{Method}
In this section, we first formulate spatial control as a unified linear inverse problem (Sec.~\ref{sec:linear_inverse}). We then introduce ProjFlow, our kinematics-aware projection sampler (Sec.~\ref{sec:projflow}), and demonstrate its use in representative applications (Sec.~\ref{sec:applications}). 

\subsection{Spatial Motion Control as a Linear Inverse Problem}
\label{sec:linear_inverse}
We unify all user-specified constraints into a single linear observation model
\begin{align}
    \vy \;=\; A \vx \;+\; \vepsilon, 
    \qquad \vepsilon \sim \mathcal{N}(\mathbf{0},\,\Sigma),
    \label{eq:obs}
\end{align}
where $\vy \in \mathbb{R}^m$ is the vector of user-specified observed measurements, $A:\mathbb{R}^{d}\!\to\!\mathbb{R}^{m}$ is a known linear operator, and $\Sigma\succeq \mathbf{0}$ is an observation noise covariance. Hard constraints are recovered as the limiting case where the corresponding rows of $\Sigma$ tend to zero variance.

Our objective is to generate a motion $\hat{\vx}$ that is consistent with the observation model~\eqref{eq:obs} while maintaining the realism encoded in the pretrained motion prior.

\subsection{Projection Sampling with Flow Matching}
\label{sec:projflow}
Given the intermediate state $\vx_t$ and the predicted velocity $v_\theta(\vx_t,t)$, as shown in Fig.~\ref{fig:projflow}, the corresponding clean endpoints can be obtained by Tweedie's formula~\cite{kim2025flowdps}
\begin{align}
\hat{\vx}_1 &=\E[\vx_1|\vx_t] = \vx_t + (1-t)v_\theta(\vx_t,t).
\end{align}
We seek the smallest clean-endpoint correction $\Delta\vx_1$ (in the metric $R\!\succ\!0$) by solving the problem
\begin{align}
\label{eq:delta_obj}
\min_{\Delta\vx_1} 
\frac12\|\Delta\vx_1\|_{R}^{2}
+
\frac12\|\vy - A(\hat{\vx}_1 + \Delta\vx_1)\|_{\Sigma^{-1}}^{2}.
\end{align}
This convex quadratic problem has a unique closed-form solution $\Delta\vx_1^{\star}$ given by
\begin{align}
\label{eq:delta_closed}
\Delta\vx_1^{\star} 
= R^{-1}A^\top\bigl(AR^{-1}A^\top + \Sigma\bigr)^{-1}\bigl(\vy - A\hat{\vx}_1\bigr).
\end{align}
Applying this to $\hat{\vx}_1$ yields the corrected clean endpoint
\begin{align}
\label{eq:x1_star}
\hat{\vx}_1^{\star} \;=\; \hat{\vx}_1 + \Delta\vx_1^{\star}.
\end{align}
We then compute the next state $\vx_{t+\Delta t}$ by adapting the stochastic recomposition step from the FlowDPS sampler~\cite{kim2025flowdps}. This step combines our corrected clean endpoint $\hat{\vx}_1^{\star}$ with a mixed version of the \emph{original} noise $\vx_0$:
\begin{align}
    \label{eq:noise-mix}
    \tilde{\vx}_0 &= \sqrt{1-\eta_t} \vx_0 + \sqrt{\eta_t} \vepsilon, \quad \vepsilon \sim \mathcal{N}(0, I) \\
    \label{eq:recomposition}
    \vx_{t+\Delta t} &= \alpha_{t+\Delta t} \hat{\vx}_1^{\star} + \sigma_{t+\Delta t} \tilde{\vx}_0,
\end{align}
where $\eta_t$ is a noise-mixing parameter, and the path coefficients are defined as $\alpha_{t+\Delta t} = t+\Delta t$ and $\sigma_{t+\Delta t} = 1 - (t+\Delta t)$.

\vspace{-1em}
\paragraph{Kinematics-aware Metric}
The choice of metric $R$ determines how we measure the size of a correction $\Delta \vx_1$ in the clean motion space. With the Euclidean metric ($R=I$), all coordinates are weighted equally, so slight changes to a few joints may appear ``small'' in terms of $\ell_2$ norm even if it breaks kinematic coherence.
We instead define smallness by coherence along the kinematic tree. The full metric $R$ for a motion $\vx \in \mathbb{R}^{d}$ is defined as
\begin{align}
\label{eq:motion-sobolev}
R = w_{\mathrm{kin}} (I_3 \otimes I_N \otimes L_{\mathrm{kin}}) + \lambda I_{d},
\end{align}
where $L_{\mathrm{kin}} \in \mathbb{R}^{J \times J}$ is the standard unnormalized graph Laplacian of the skeletal topology. It is constructed from the skeleton's adjacency matrix $A_{\text{kin}}$ (where $(A_{\text{kin}})_{j_1j_2} = 1$ if joint $j_1$ and $j_2$ are connected) as $L_{\text{kin}} = D_{\text{kin}} - A_{\text{kin}}$, with the diagonal degree matrix $D_{\text{kin}} = \text{diag}(A_{\text{kin}} \mathbf{1})$. $I_k$ is the $k \times k$ identity matrix, $w_{\mathrm{kin}}$ is a scalar weight for the kinematic term, and $\lambda > 0$ is a weight for the identity term, which ensures $R$ is strictly positive definite and invertible. This metric is applied independently to each of the $x, y, \text{and } z$ spatial dimensions via the $I_3$ term.

This metric makes the intended measurement of ``small'' explicit (i) discrepancies across \emph{adjacent joints} are strongly penalized by the kinematic term $w_{\mathrm{kin}} L_{\mathrm{kin}}$, while joints that are not directly connected in the kinematic tree incur little coupling, reflecting the skeletal topology. (ii) The identity term $\lambda I$ adds a baseline $\ell_2$ penalty to directions, which are per-frame global translations that are not penalized by the kinematic component. This penalty regularizes these otherwise unconstrained modes and ensures that the full metric $R$ is strictly positive definite.

\subsection{Spatial Control with ProjFlow}
\label{sec:applications}
We illustrate ProjFlow in practice through two representative spatial control applications: motion inpainting and 2D-to-3D lifting. Other extensions, such as motion loop closure and relative body part control shown in Fig.~\ref{fig:teaser}, are formulated in the supplementary material.

\subsubsection{Application I: Motion Inpainting via Masked Pseudo-Observations}
\begin{figure}[t]
\centering
\includegraphics[width=0.9\linewidth]{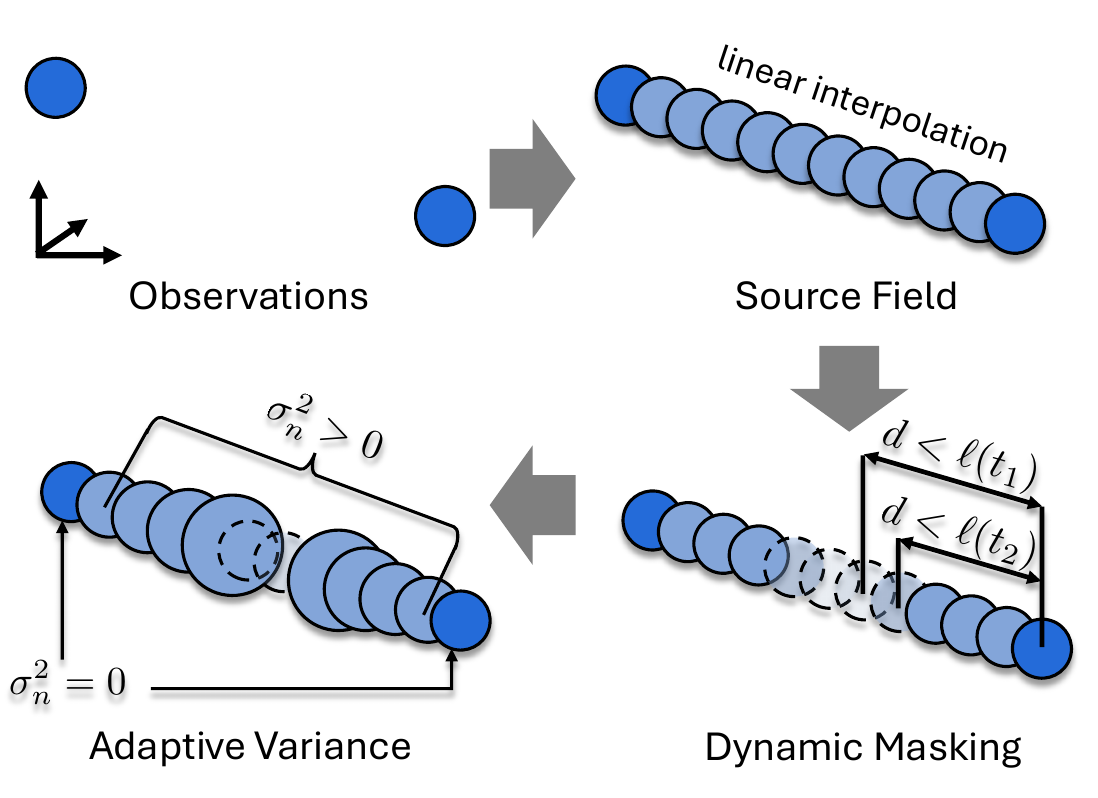}
\caption{\textbf{Pseudo-observations for motion inpainting.} Sparse observations are interpolated to guide intermediate frames. This guidance is controlled by two mechanisms: Dynamic Masking activates a time-scheduled neighborhood, and Adaptive Variance treats original observations as hard constraints and the interpolated guides as soft constraints.}
\label{fig:motion_inpaint}
\end{figure}

\noindent\textbf{Plain Masking.}
We cast inpainting as recovering the full motion vector $\vx\in\mathbb{R}^d$ from sparse hard observations, such as keyframe joint locations provided by users. Let $M_{\mathrm{obs}}\in\{0,1\}^{d\times d}$ be a diagonal mask selecting observed coordinates, and $\vy_{\mathrm{obs}}\in\mathbb{R}^d$ store their values (zeros elsewhere). The hard‑constraint model is
\begin{align}
\label{eq:inpaint_basic}
\vy_{\mathrm{obs}} \;=\; M_{\mathrm{obs}}\vx.
\end{align}

\noindent\textbf{Time‑varying Pseudo‑observations.}
When these hard observations are sparse, the model provides insufficient guidance. We therefore introduce “soft” pseudo-observations $\vy_{\mathrm{src}}$, created via per-joint linear interpolation, to provide denser guidance.
However, these pseudo-observations from linear interpolation are not always reliable. We want the variance to be high (i.e., trust is low) in two cases
(i) As sampling progresses ($t \to 1$), we trust the model's own prediction $\hat{\vx}_1$ more.
(ii) Where motion curvature is high, linear interpolation is a poor estimate.

We combine these soft guides with the hard observations $\vy_{\mathrm{obs}}$ to formulate a time-varying linear inverse problem at each sampling step $t$
\begin{align}
\vy^{(t)} \;=\; M^{(t)}\vx + \vepsilon^{(t)},\qquad
\vepsilon^{(t)} \sim \mathcal{N}\!\big(0,\Sigma^{(t)}\big),
\end{align}
where $M_{\mathrm{aug}}^{(t)}$ is a diagonal matrix activating pseudo-observations within a temporal neighbourhood of hard constraints, but explicitly excluding the hard constraints themselves. The combined mask is the union of these disjoint sets, $M^{(t)} = M_{\mathrm{obs}} + M_{\mathrm{aug}}^{(t)}$. The target observation is $\vy^{(t)} = \vy_{\mathrm{obs}} + M_{\mathrm{aug}}^{(t)}\,\vy_{\mathrm{src}}$. The diagonal covariance $\Sigma^{(t)}=\mathrm{diag}(\sigma_1^2(t),\ldots,\sigma_d^2(t))$ assigns an adaptive, non-zero variance $\sigma_i^2(t) > 0$ to the active pseudo-observations based on their reliability. The actual observations are treated as exact linear equalities. 

\noindent\textbf{Dynamic Masking.}
The temporal neighbourhood of pseudo-observations (Fig.\ref{fig:motion_inpaint}, Dynamic Masking) shrinks linearly in time. This mechanism gradually phases out the soft pseudo-observations, leaving only the hard constraints active as $t \to 1$. We define this shrinking radius $\ell(t)$ as
\begin{align}
\ell(t) \;=\; (1-t)\,\ell_{\max} + t\,\ell_{\min}.
\end{align}
A frame's pseudo-observations are activated only if the temporal distance to its nearest hard observation is less than this radius $\ell(t)$.

\noindent\textbf{Adaptive Variance.}
We control the reliability of the pseudo-observations by setting their variance $\sigma_i^2(t)$ (Fig.\ref{fig:motion_inpaint}, Adaptive Variance).
We model the trust level with a frame-wise score $\tilde{\pi}_n^{(t)}$
\begin{align}
\tilde{\pi}_n^{(t)} \;=\; \tau(t)\,\frac{c_0}{1+\lambda_s\,(s_n(\hat{\vx}_1)/s_{\text{med}})^p}
\end{align}
where $c_0$, $\lambda_s$, and $p$ are hyperparameters controlling the adaptive strength.
This score combines a global time-decay term,
\begin{align}
\tau(t) \;=\; \tau_{\min}+(1-\tau_{\min})(1-t),
\end{align}
where $\tau_{\min}$ is a hyperparameter, with a local curvature penalty $s_n(\hat{\vx}_1)$, defined as
\begin{align}
s_n(\hat{\vx}_1) \;=\; \|(\hat{\vx}_1)_{n+1}-2(\hat{\vx}_1)_n+(\hat{\vx}_1)_{n-1}\|_R.
\end{align}
Here, $s_{\text{med}}$ is the median curvature $s_n(\hat{\vx}_1)$ across the sequence, used for robust normalization. As time $t$ increases or curvature $s_n$ increases, the trust score $\tilde{\pi}_n^{(t)}$ decreases. We clip this score to get a frame-level base target $\pi_n^{(t)} = \mathrm{clip}\!\big(\,\tilde{\pi}_n^{(t)},\ \pi_{\min},\ \pi_{\max}\,\big)$. This base score is then modulated per-joint based on the properties of the kinematic metric to yield the final per-element score $\pi_i$. This $\pi_i$ is used to compute the variance $\sigma_i^2(t)$ for the active pseudo-observation via the relation $\pi_i = r_i/(r_i+\sigma_i^2(t))$. Solving for the variance gives
\begin{align}
\sigma_i^2(t) \;=\; r_i \frac{1 - \pi_i}{\pi_i}
\end{align}
where $r_i = [\mathrm{diag}(R^{-1})]_i$ is the $i$-th diagonal element of the inverse kinematic metric. Hard observations always maintain zero variance ($\sigma_i^2=0$).

\begin{figure*}[t]
\centering
\includegraphics[width=\linewidth]{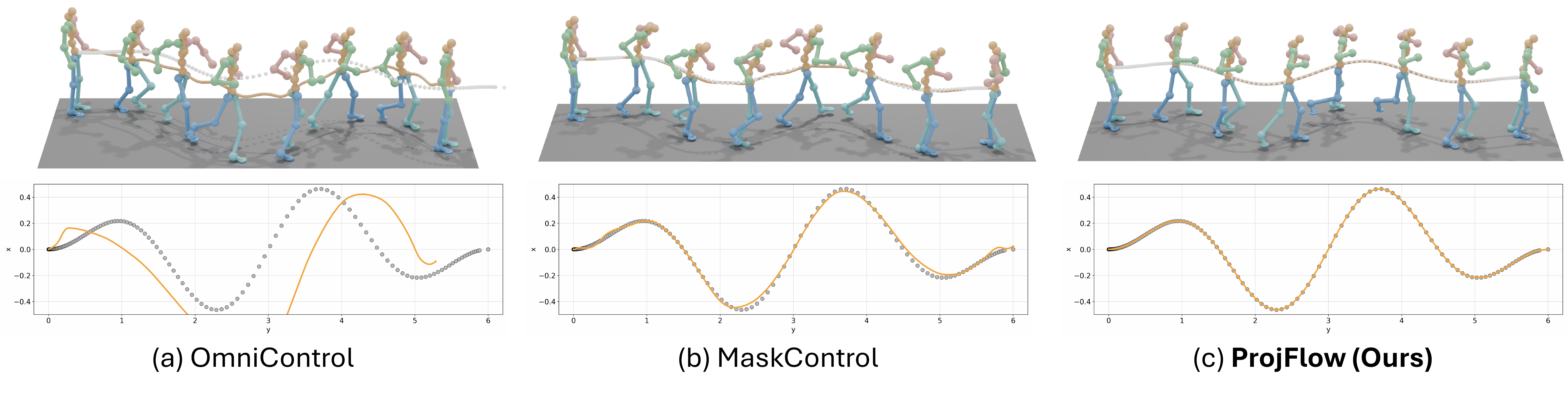}
\vspace{-1em}
    \caption{\textbf{Text-conditioned pelvis-trajectory control.} Given the prompt ``\textit{a person runs forward in an S-shaped path}'' and a pelvis control signal, we compare OmniControl~\citep{xie2024omnicontrol}, MaskControl~\citep{pinyoanuntapong2024controlmm}, and ProjFlow (ours). The rendered motions and the trajectory plots both visualize the generated pelvis trajectory ({\color{motion}orange}) overlaid on the target control signal ({\color{hint} gray dotted line}).
    }
\label{fig:traj_control}
\end{figure*}

\subsubsection{Application II: 2D-to-3D Lifting via Linear Projection Measurements}
\label{sec:lifting}
The 2D-to-3D motion lifting task can also be expressed as a linear inverse problem. In this setting, we assume noise-free hard constraints, so the model simplifies to $\vy = A\,\vx$. The operator $A$ maps the vectorized 3D motion sequence $\vx$ to stacked 2D joint coordinates. This operator is constructed in two steps. First, we define a full projection operator $A_{\mathrm{full}}$ that maps all 3D joints at all frames to 2D. It does this by stacking the standard linear orthographic projection,
\begin{align}
\vy_{n,j} = sPR_{\text{cam}}\vx_{n,j},
\end{align}
for every frame $n$ and joint $j$, where $s$ is a fixed scale factor, $P=[\,1~0~0;~0~1~0\,]$ is the orthographic projection matrix, and $R_{\text{cam}}\!\in\!\mathrm{SO}(3)$ is the camera rotation. Both $s$ and $R_{\text{cam}}$ are assumed to be known for each sequence.

Second, we define a binary selection operator $M$ that filters the rows of $A_{\mathrm{full}}$ to match the user's specific inputs (e.g., all joints at frame 0 and a subset of joints for $n > 0$). $M$ is constructed to select only these corresponding rows.
The final measurement operator $A$ is therefore defined as
\begin{align}
A = M\,A_{\mathrm{full}}.
\end{align}

%% file: sec/5_experiment.tex
\section{Experiments}
In this section, we evaluate the performance of ProjFlow, comparing it to previous task-specific/zero-shot methods.

\subsection{Experimental Setup}
\noindent\textbf{Datasets.} We experiment on the popular HumanML3D~\cite{guo2022humanml3d} dataset which contains 14,646 text-annotated human motion sequences from AMASS~\cite{mahmood2019amass} and HumanAct12~\cite{guo2020action2motion} datasets. 

\noindent\textbf{Evaluation Protocol.}
We adopt the pretrained ACMDM-S-PS22~\citep{meng2025acmdm} as our base flow-matching model for all experiments and primarily follow the protocol of \citet{meng2025rethinking}. For spatial control experiments, we follow the OmniControl~\citep{xie2024omnicontrol} evaluation protocol, which varies the density of control signals across five settings (1, 2, 5, 49, and 196 keyframes), and report the mean of each control metric across these densities to assess robustness to sparsity. 

For the 2D-to-3D task, we follow the Sketch2Anim~\citep{zhong2025sketch2anim} protocol, which defines camera parameters including $\text{pitch}\in[0^\circ,30^\circ]$, $\text{yaw}\in[-45^\circ,45^\circ]$, $\text{roll}=0^\circ$, and $s\in[0.8,1.2]$. We evaluate under this known orthographic camera at inference time.

\noindent\textbf{Evaluation Metrics.}
To assess generation quality and text alignment, we report \textit{FID} for distribution similarity, \textit{R-Precision (Top-1/2/3)} and \textit{Matching Score} for semantic retrieval accuracy between motion and text embeddings, \textit{Diversity} for motion diversity. For spatial control tasks, we evaluate accuracy using \textit{Trajectory Error}, \textit{Location Error}, and \textit{Average Error}, which measure deviations from target keyframes at trajectory, keyframe, and mean distance levels, respectively. Physical plausibility is assessed via the \textit{Foot Skating Ratio}. 

For the 2D-to-3D reconstruction task, in addition to the above metrics, we report \textit{MPJPE‑2D} and \textit{Avg.~Err.‑2D}. These metrics evaluate constraint satisfaction by projecting the generated 3D motion back into 2D and quantifying the mean error against the target 2D joint coordinates, following the protocol of Sketch2Anim~\citep{zhong2025sketch2anim}.

\input{table/control_67dim}

\subsection{Results}
\subsubsection{Motion Inpainting with Trajectory Control}
\noindent\textbf{Quantitative Performance.}
ProjFlow is the only zero-shot method that achieves \emph{exact} constraint satisfaction ($0.0000$ on trajectory/location/average errors) while also attaining the best realism among zero-shot baselines. As shown in Table~\ref{tab:control_noAITS}, its FID is lower than DNO(ACMDM-S-PS22+DNO)~\citep{karunratanakul2024dno} for both pelvis control and all joints, which indicates that ProjFlow can eliminate the small residual violations that remain for guidance/noise-optimization methods.

Compared to models that require additional training, as shown in Table~\ref{tab:control_noAITS}, ProjFlow stays in a similar realism band while remaining training-free and achieving \emph{exact} constraint satisfaction. For example, 
MaskControl~\citep{pinyoanuntapong2024controlmm} reaches a lower FID but still leaves a non-zero average error ($0.0093$), whereas ProjFlow maintains all control errors at $0.0000$. The same tendency is observed in other training-based controllers such as OmniControl~\citep{xie2024omnicontrol}. Even when the same base model is additionally trained with a ControlNet branch (ACMDM-S-PS22+CtrlNet), the constraints are still not fully satisfied, despite a slightly improved FID of 0.067. In contrast, ProjFlow achieves exact constraint satisfaction without any retraining.

\noindent\textbf{Qualitative Analysis.}
Fig.~\ref{fig:traj_control} compares the generated motions from OmniControl~\citep{xie2024omnicontrol}, MaskControl~\citep{pinyoanuntapong2024controlmm}, and ProjFlow. OmniControl~\citep{xie2024omnicontrol} captures the overall S-shaped tendency of the target path but deviates significantly along the curve, especially near the bends. MaskControl~\citep{pinyoanuntapong2024controlmm} uses a ControlNet-style branch and additionally performs inference-time optimization, which further reduces this deviation. However, close inspection of the overlaid trajectories still reveals slight mismatches between the generated and target paths. By contrast, ProjFlow aligns the generated pelvis trajectory with the target markers essentially exactly across the entire S-shaped path while preserving natural full-body motion.

\begin{figure*}[t]
\centering
\includegraphics[width=\linewidth]{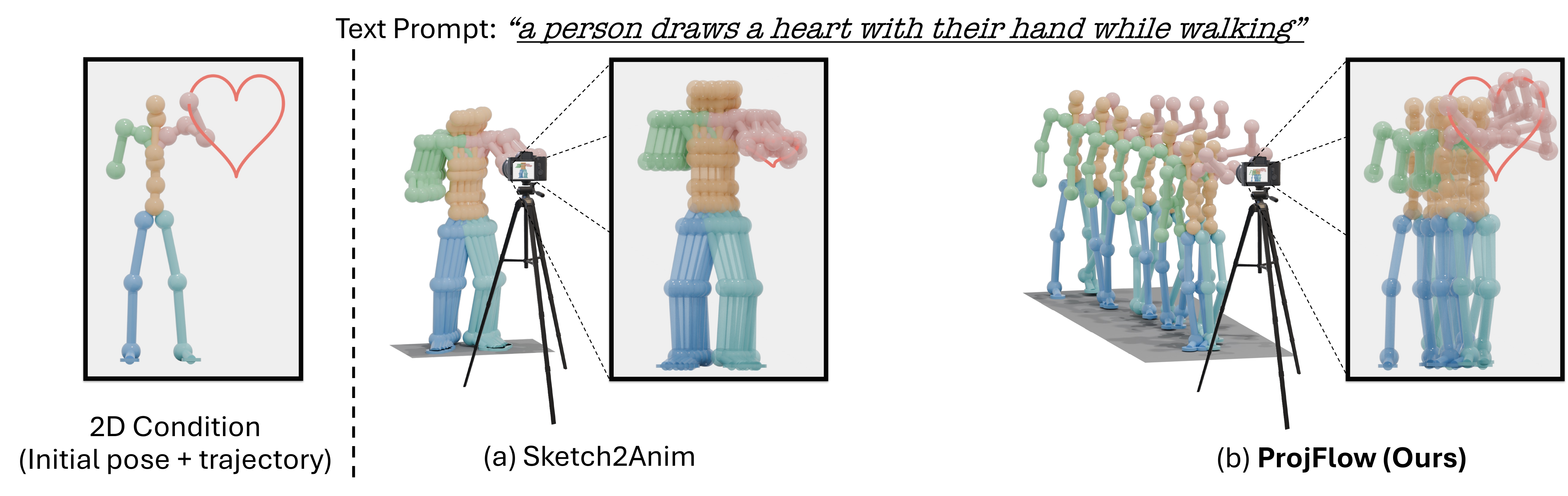}
\vspace{-2em}
\caption{\textbf{2D-to-3D hand-trajectory lifting with text conditioning.} The input condition includes the text prompt ``\textit{a person draws a heart with their hand while walking},'' an initial 2D keypose, and a left-wrist 2D trajectory shaped like a heart. Sketch2Anim~\citep{zhong2025sketch2anim} fails to reproduce the heart path precisely, the shape collapses, and the subject does not exhibit walking motion. In contrast, ProjFlow follows the heart-shaped wrist trajectory accurately while maintaining a natural walking motion throughout the sequence.}
\label{fig:2d3d_heart}
\end{figure*}
\input{table/camera}
\subsubsection{2D-to-3D Reconstruction}
\noindent\textbf{Quantitative Performance.}
As shown in Table~\ref{tab:compare_with_baseline}, ProjFlow achieves superior motion naturalness, attaining a lower FID than the state-of-the-art method Sketch2Anim~\citep{zhong2025sketch2anim} under both Average and Cross evaluation protocols. For constraint satisfaction, ProjFlow enforces the 2D constraints \emph{exactly} to the numerical precision (MPJPE-2D $=\mathbf{0.000}$) while Sketch2Anim~\citep{zhong2025sketch2anim} still exhibits residual reprojection errors.

\noindent\textbf{Qualitative Analysis.}
Fig.~\ref{fig:2d3d_heart} shows a qualitative example of the 2D-to-3D lifting task. The goal is to generate a 3D motion that follows the given 2D heart-shaped wrist trajectory and the given initial 2D keypose, while simultaneously "walking" as specified by the text prompt.

ProjFlow succeeds in following the 2D heart trajectory exactly at every frame while keeping the other joints engaged in a natural walking motion. The legs and torso continue to produce smooth, coordinated gait cycles as the left wrist draws the heart shape in the image plane. In contrast, Sketch2Anim~\citep{zhong2025sketch2anim} fails to preserve the heart shape, and the trajectory collapses into a distorted loop. The character also primarily remains in place, only moving the arm without translating forward, indicating that the intended instruction to walk is not realized.

\subsubsection{Ablation Study}
We analyze the contribution of ProjFlow's three key components on the motion inpainting task in Table~\ref{tab:ablation_control_67}. First, replacing our kinematics-aware metric with a standard Euclidean metric severely degrades motion realism, causing a significant degradation in FID. This confirms that propagating corrections coherently along the skeleton is critical. Second, removing the stochastic recomposition step ($\eta_t=0$) and deterministically recomposing the state also drastically harms quality and diversity. This highlights the importance of noise mixing for staying on the learned motion manifold. Third, for the inpainting task, reverting to a "Plain masking" approach without our pseudo-observation significantly worsens realism. These results validate that while all variants maintain exact constraint satisfaction, all three proposed components are essential for generating natural and realistic motion.

\input{table/ablation_control}

%% file: table/control_67dim.tex
\begin{table*}[t]
    \centering
    \caption{\textbf{Quantitative text-conditioned motion generation with spatial control signals and upper-body editing on HumanML3D\cite{guo2022humanml3d}.} In the first section, methods are trained and evaluated solely on pelvis controls. In the middle section, methods are trained on all joints and evaluated separately on each controlled joint. Only average results are reported for brevity. We include details in the supplementary material. The last section presents upper-body editing results. \textbf{bold} face / \underline{underline} indicates the best/2\textsuperscript{nd} results.}
    \vspace{-0.5em}
    \renewcommand{\arraystretch}{0.92}
    \resizebox{1\linewidth}{!}{
    \begin{tabular}{clcccccccc}
    \toprule
       \multirowcell{2}{Controlling\\ Joint} & \multirow{2}{*}{Methods} & \multirow{2}{*}{Zero-shot?}& \multirow{2}{*}{FID$\downarrow$}& R-Precision & \multirow{2}{*}{Diversity$\rightarrow$} & Foot Skating& \multirow{2}{*}{Traj. err.$\downarrow$} & \multirow{2}{*}{Loc. err.$\downarrow$} & \multirow{2}{*}{Avg. err.$\downarrow$}\\
    ~ & && & Top 3 & & Ratio.$\downarrow$ & & & \\
    \midrule
    &GT &- &$0.000$ &$0.795$ &$10.455$ &-&$0.000$ &$0.000$ &$0.000$\\
    \midrule
    \multirow{9}{*}{\parbox{2.2cm}{\centering \textbf{Pelvis}}}
    &MDM~\citep{tevet2023human} &\ding{51} &$1.792$ &$0.673$ &$9.131$ &$0.1019$ &$0.4022$ &$0.3076$ &$0.5959$\\
    &PriorMDM~\citep{shafir2024priormdm} &\ding{55} &$0.393$ &$0.707$ &$9.847$ &$0.0897$&$0.3457$ &$0.2132$ &$0.4417$\\
    &GMD~\citep{karunratanakul2023gmd} &\ding{51} &$0.238$ &$0.763$ &$10.011$ &$0.1009$&$0.0931$ &$0.0321$ &$0.1439$\\
    &OmniControl~\citep{xie2024omnicontrol} &\ding{55} &$0.081$ &$0.789$ &$10.323$  &$\underline{0.0547}$ &$0.0387$ &$0.0096$ &$0.0338$\\
    &MotionLCM V2+CtrlNet~\citep{dai2024motionlcm} &\ding{55} &$3.978$ &$0.738$ &$9.249$ &$0.0901$ &$0.1080$ &$0.0581$&$0.1386$ \\
    &MaskControl~\citep{pinyoanuntapong2024controlmm} &\ding{55}&$\mathbf{0.066}$ &$0.799$ &$\mathbf{10.474}$ &$\mathbf{0.0543}$ &$\mathbf{0.0000}$ &$\mathbf{0.0000}$& $0.0093$\\
    &ACMDM\text{-}S\text{-}PS22+CtrlNet~\citep{meng2025acmdm} &\ding{55} &$\underline{0.067}$ &$\mathbf{0.805}$ &$\underline{10.481}$ &$0.0591$ &$0.0075$ &$0.0010$ 
    &$0.0100$\\
    &ACMDM\text{-}S\text{-}PS22+DNO~\citep{karunratanakul2024dno} &\ding{51} &$0.151$ &$\underline{0.802}$ &$-$ &$0.0610$ &$\underline{0.0027}$ &$\underline{0.0002}$ 
    &$\underline{0.0089}$\\
    &\textbf{ACMDM\text{-}S\text{-}PS22+ProjFlow} &\ding{51} &$0.107$ &$0.784$ &$10.644$ &$0.0629$ &$\mathbf{0.0000}$ &$\mathbf{0.0000}$ &$\mathbf{0.0000}$\\

    \midrule

    \multirow{6}{*}{\parbox{2.2cm}{\centering \textbf{All Joints\\(Average)}}}
    &OmniControl~\citep{xie2024omnicontrol} &\ding{55} &$0.126$ &$0.792$ &$\underline{10.276}$ & $0.0608$&$0.0617$ &$0.0107$ &$0.0404$\\
    &MotionLCM V2+CtrlNet~\citep{dai2024motionlcm} &\ding{55} &$4.504$ &$0.715$ &$9.230$ &0.1119 &$0.2740$ &$0.1315$ &$0.2464$\\
    &MaskControl~\citep{pinyoanuntapong2024controlmm} &\ding{55}&$\underline{0.095}$ &$0.795$ &$10.159$ &$\mathbf{0.0545}$ &$\mathbf{0.0000}$ &$\mathbf{0.0000}$& $\underline{0.0065}$\\
    &ACMDM\text{-}S\text{-}PS22+CtrlNet~\citep{meng2025acmdm} &\ding{55} &$\mathbf{0.070}$ &$\mathbf{0.803}$ &$\mathbf{10.526}$ &$\underline{0.0596}$ &$0.0117$ &$0.0019$ &$0.0197$\\
    &ACMDM\text{-}S\text{-}PS22+DNO~\citep{karunratanakul2024dno} &\ding{51} &$0.147$ &$\underline{0.800}$ &$-$ &$0.0600$ &$\underline{0.0034}$ &$\underline{0.0003}$ 
    &$0.0121$\\
    &\textbf{ACMDM\text{-}S\text{-}PS22+ProjFlow} &\ding{51} &$0.097$ &$0.779$ &$10.651$ &$0.0603$ &$\mathbf{0.0000}$ &$\mathbf{0.0000}$ &$\mathbf{0.0000}$\\

    \midrule
    \midrule
    & \multirow{2}{*}{Methods} & \multirow{2}{*}{Zero-shot?}& \multirow{2}{*}{FID$\downarrow$}& R-Precision & R-Precision  & R-Precision & \multirow{2}{*}{Matching$\downarrow$} & \multirow{2}{*}{Diversity$\rightarrow$} & \multirow{2}{*}{$-$}\\
    ~ & && & Top 1 &Top 2 & Top 3& & \\
    \midrule
    \multirow{6}{*}{\parbox{2.2cm}{\centering \textbf{Upper-Body\\Edit}}} 
    &MDM~\citep{tevet2023human} &\ding{51}&$1.918$ &$0.359$&$0.556$&$0.654$&$4.793$&$9.210$&$-$\\
    &OmniControl~\citep{xie2024omnicontrol} &\ding{55}&$0.909$ &$0.428$&$0.614$&$0.722$&$3.694$&$10.207$&$-$\\
    &MotionLCM V2+CtrlNet~\citep{dai2024motionlcm} &\ding{55}&$3.922$&$0.404$ &$0.592$&$0.692$&$5.610$&$9.309$&$-$\\
    &MaskControl~\citep{pinyoanuntapong2024controlmm} &\ding{55}&$\mathbf{0.066}$ &$\underline{0.501}$ &$\underline{0.695}$ &$\underline{0.794}$ &$\underline{3.227}$& $10.159$ & $-$\\
    &ACMDM\text{-}S\text{-}PS22+CtrlNet~\citep{meng2025acmdm} &\ding{55} &$\underline{0.076}$ &$\mathbf{0.532}$&$\mathbf{0.719}$&$\mathbf{0.820}$&$\mathbf{3.098}$&$\underline{10.586}$&$-$\\
    &\textbf{ACMDM\text{-}S\text{-}PS22+ProjFlow}&\ding{51} &$0.087$ &$\underline{0.501}$ &$0.690$ &$0.787$ &$3.319$ &$\mathbf{10.571}$ &$-$\\
    \bottomrule
    \end{tabular}}
    \label{tab:control_noAITS}
\end{table*}

%% file: table/camera.tex
\begin{table*}[t]
\centering
\caption{\textbf{Quantitative analysis of ProjFlow and three baseline models proposed in Sketch2Anim~\citep{zhong2025sketch2anim} on the HumanML3D~\citep{guo2022humanml3d}.} Evaluation metrics on motion realism, control accuracy, and text-motion match are presented. Following OmniControl~\citep{xie2024omnicontrol}, we report both the average error of all joints (Average) and their random combination (Cross). \textbf{bold} face / \underline{underline} indicates the best/2\textsuperscript{nd} results.
}
\vspace{-0.5em}
\label{tab:compare_with_baseline}
\resizebox{\textwidth}{!}{
\begin{tabular}{cccccccccc}
\toprule[0.25ex]
\multirow{2}{*}{Condition} & \multirow{2}{*}{Method} & \multicolumn{2}{c}{Realism} & \multicolumn{4}{c}{Control Accuracy} & \multicolumn{2}{c}{Text-Motion Matching} \\
\cmidrule(lr){3-4} \cmidrule(lr){5-8} \cmidrule(lr){9-10}
 &  & FID $\downarrow$ & Foot Skating $\downarrow$ & MPJPE-2D $\downarrow$ & MPJPE-3D $\downarrow$ & Avg. Err.-2D $\downarrow$ & Avg. Err.-3D $\downarrow$ & Matching$\downarrow$ & R-precision (Top-3) $\uparrow$ \\
\midrule
\multirow{5}{*}{Average} 
 & Motion Retrieval & 0.690 & \textbf{0.064} & 0.057 & 0.076 & 0.290 & 0.410 & 4.060 & 0.640 \\
 & Lift-and-Control & 0.979 & \underline{0.089} & 0.054 & 0.071 & 0.261 & 0.340 & \underline{3.297} & \underline{0.752} \\
 & Direct 2D-to-Motion & 2.553 & 0.112 & 0.040 & 0.055 & 0.193 & \underline{0.275} & 3.723 & 0.687 \\
 & Sketch2Anim~\citep{zhong2025sketch2anim} &  \underline{0.525} & 0.103 & \underline{0.036} & \underline{0.048} & \underline{0.087} & \textbf{0.134} & \textbf{3.077} & \textbf{0.802} \\
& \textbf{ACMDM-S-PS22+ProjFlow}  &  \textbf{0.349} & 0.146 & \textbf{0.000} & \textbf{0.042} & \textbf{0.000} & 0.331 & 3.363 & 0.748\\
\midrule
\multirow{5}{*}{Cross} 
 & Motion Retrieval & \textbf{0.103} & \textbf{0.067} & 0.055 & 0.073 & 0.307 & 0.423 & 3.405 & 0.724 \\
 & Lift-and-Control & 0.738 & \underline{0.101} & 0.051 & 0.067 & 0.209 & 0.283 & \underline{3.135} & \underline{0.778} \\
 & Direct 2D-to-Motion & 2.310 & 0.123 & 0.040 & 0.056 & 0.189 & \underline{0.266} & 3.606 & 0.709 \\
 & Sketch2Anim~\citep{zhong2025sketch2anim}  & 0.577& 0.102 & \underline{0.033} & \underline{0.046} & \underline{0.079} & \textbf{0.132} & \textbf{3.042} & \textbf{0.796} \\
& \textbf{ACMDM-S-PS22+ProjFlow} &  \underline{0.168} & 0.139 & \textbf{0.000} & \textbf{0.037} & \textbf{0.000} & 0.298 & 3.259 & 0.764 \\  
\bottomrule[0.25ex]
\end{tabular}
}
\vspace{-1.4em}
\end{table*}

%% file: table/ablation_control.tex
\begin{table}[t]
    \centering
    \small
    \renewcommand{\arraystretch}{1.2}
    \caption{Ablation studies of ProjFlow.}
    \vspace{-1em}
    \resizebox{\linewidth}{!}{
    \begin{tabular}{@{}llccccccc@{}}
    \toprule
    \textbf{Variant} & FID$\downarrow$ & R-Prec. & Div.$\rightarrow$ & Foot$\downarrow$ & Traj.$\downarrow$ & Loc.$\downarrow$ & Avg.$\downarrow$ \\
    \midrule
    ProjFlow (Full) & 0.097 & 0.779 & 10.651 & 0.0603 & 0.0000 & 0.0000 & 0.0000 \\
    \midrule
    Euclid. ($R{=}I$)  & 1.152 & 0.740 & 10.107 & 0.0595 & 0.0000 & 0.0000 & 0.0000 \\
    No noise ($\eta_t{=}0$) & 3.429 & 0.707 & 9.307 & 0.0863 & 0.0000 & 0.0000 & 0.0000 \\
    Plain masking & 0.880 & 0.748 & 10.187 & 0.0632 & 0.0000 & 0.0000 & 0.0000 \\
    \bottomrule
    \end{tabular}}
    \label{tab:ablation_control_67}
    \vspace{-0.5em}
\end{table}

%% file: sec/6_limitations.tex
\section{Limitations}
While ProjFlow offers exact satisfaction of linear spatial constraints in a training-free manner, it is fundamentally limited to constraints that can be formulated as linear inverse problems. Our framework, in its current form, cannot natively handle more complex non-linear constraints. Examples of such constraints include inequalities such as keeping a joint above a certain plane. Extending the closed-form projection to these more expressive, non-linear scenarios is a challenging but important direction for future work. 

%% file: sec/7_conclusion.tex
\section{Conclusion}
In this paper, we presented \textbf{ProjFlow}, a zero-shot projection sampler for flow-matching models that achieves exact spatial motion control. Our method unifies diverse animation tasks, such as trajectory following and 2D-to-3D lifting, by formulating them as linear inverse problems. The sampler projects the clean motion estimate onto the linear constraint set at each ODE step. This projection employs a novel kinematics-aware metric that respects skeletal topology to maintain motion naturalness. ProjFlow successfully enforces hard constraints exactly without requiring any task-specific retraining or iterative optimization. Experiments on motion inpainting and 2D-to-3D reconstruction show that our framework matches the realism of training-based methods while guaranteeing exact constraint satisfaction. ProjFlow provides a practical route for interactive and precise motion authoring.

%% file: sec/suppl.tex
\clearpage
\setcounter{page}{1}
\maketitlesupplementary

\noindent This supplementary material is organized as follows:
\begin{itemize}
    \item Section~\ref{sec:supp_analysis}: Analytical view of ProjFlow.
    \item Section~\ref{sec:supp_methods}: Additional method details.
    \item Section~\ref{sec:supp_impl}: Implementation details.
    \item Section~\ref{sec:supp_results}: Additional quantitative results.
    \item Section~\ref{sec:supp_results_video}: Additional qualitative results.
\end{itemize}

\section{Analytical View of ProjFlow}
\label{sec:supp_analysis}
Using the notation in Table~\ref{tab:supp_notation}, we provide an analytical interpretation of the ProjFlow update, including its relation to DDNM and a MAP view. In what follows, ``PSD'' and ``PD'' denote positive semidefinite and positive definite matrices, respectively.
\vspace{-1em}
\begin{table}[ht]
\centering
\small
\caption{\textbf{Notation used in the supplementary derivations.}}
\vspace{-1em}
\label{tab:supp_notation}
\begin{tabular}{lll}
\hline
\textbf{Symbol} & \textbf{Type / shape} & \textbf{Note} \\
\hline
$A$           & $\mathbb{R}^{m\times d}$   & linear operator \\
$\vy$         & $\mathbb{R}^{m}$           & measurements \\
$\vx_1$       & $\mathbb{R}^{d}$           & clean motion endpoint \\
$\hat{\vx}_1$ & $\mathbb{R}^{d}$           & estimate of $\vx_1$ \\
$\Sigma$      & $\mathbb{R}^{m\times m}$   & PSD covariance \\
$R$           & $\mathbb{R}^{d\times d}$   & PD metric / precision \\
\hline
\end{tabular}
\end{table}
\vspace{-0.4em}

\subsection{Recovery of DDNM under Euclidean Metric and Noiseless Observation}
DDNM~\cite{wang2022ddnm} solves the linear inverse problem $y = A \vx$ by decomposing
$\mathbb{R}^d$ into the range and null space of $A$. Given a clean-endpoint estimate
$\hat{\vx}_1$, it keeps the range-space component consistent with the measurements and
fills the null space with $\hat{\vx}_1$:
\begin{align}
    \hat{\vx}_1^\star
    = A^\dagger \vy + (I - A^\dagger A)\hat{\vx}_1,
\end{align}
where $A^\dagger$ is the Moore--Penrose pseudoinverse of $A$.

ProjFlow, in contrast, updates the clean-endpoint estimate via
\begin{align}
    \hat{\vx}_1^\star
    = \hat{\vx}_1 + R^{-1}A^\top\bigl(AR^{-1}A^\top + \Sigma\bigr)^{-1}
      \bigl(\vy - A\hat{\vx}_1\bigr).
\end{align}
Specializing to the Euclidean metric $R = I$ and the noise-free limit $\Sigma \to 0$,
and assuming that $A$ has full row rank so that $AA^\top$ is invertible, we obtain
\begin{align}
 \hat{\vx}_1^\star
&= \hat{\vx}_1 + A^\top\bigl(AA^\top\bigr)^{-1}\bigl(\vy - A\hat{\vx}_1\bigr) \\
&= \hat{\vx}_1 + A^\dagger \vy - A^\dagger A \hat{\vx}_1 \\
&= A^\dagger \vy + \bigl(I - A^\dagger A\bigr)\hat{\vx}_1,
\end{align}
which coincides exactly with the DDNM update above. Thus, DDNM is recovered as a
special case of ProjFlow in the Euclidean, noiseless setting.

\subsection{ProjFlow as MAP Estimation}
ProjFlow's projection step can also be interpreted as computing a
maximum-a-posteriori (MAP) estimate in a linear--Gaussian model.
We treat the clean-endpoint estimate $\hat{\vx}_1$ from Tweedie's
formula as the mean of a Gaussian prior
\begin{align}
p(\vx_1)
\;=\;
\mathcal{N}\!\bigl(\vx_1 \mid \hat{\vx}_1,\; R^{-1}\bigr),
\end{align}
where $R \succ 0$ is the precision matrix and $R^{-1}$ is the
corresponding covariance.

For the Euclidean metric $R = I$, this prior is an isotropic Gaussian
centered at $\hat{\vx}_1$, penalizing all directions equally. With the
kinematics-aware metric $R$, the structure is instead governed by the
skeletal Laplacian $L_{\text{kin}}$: directions that create large
differences between adjacent joints (skeletally incoherent motion) have
small variance, while coordinated joint motions have larger variance.
Geometrically, this yields a highly anisotropic ellipsoidal prior that
favors kinematically coherent corrections.

The linear observation model is
\begin{align}
\vy &= A\vx_1 + \vepsilon,
\qquad
\vepsilon \sim \mathcal{N}(\mathbf{0},\,\Sigma)
\\[0.25em]
&\Longleftrightarrow\;
p(\vy \mid \vx_1)
=
\mathcal{N}\!\bigl(\vy \mid A\vx_1,\,\Sigma\bigr).
\end{align}
Combining this likelihood with the prior yields a Gaussian posterior
\begin{align}
p(\vx_1 \mid \vy)\;\propto\;
\exp\!\Bigl(
-\tfrac12\|\vx_1-\hat{\vx}_1\|_{R}^{2}
-\tfrac12\|\vy-A\vx_1\|_{\Sigma^{-1}}^{2}
\Bigr).
\end{align}
The MAP estimate $\vx_1^{\text{MAP}}$ maximizes this posterior, or
equivalently minimizes the negative log-posterior:
\begin{align}
\vx_1^{\text{MAP}}
&= \argmin_{\vx_1}
\Bigl(
\|\vx_1-\hat{\vx}_1\|_{R}^{2}
+\|\vy-A\vx_1\|_{\Sigma^{-1}}^{2}
\Bigr).
\end{align}
Taking the gradient with respect to $\vx_1$ and setting it to zero gives
the normal equations
\begin{align}
\bigl(R + A^\top \Sigma^{-1} A\bigr)\vx_1
= R\hat{\vx}_1 + A^\top \Sigma^{-1}\vy,
\end{align}
so that
\begin{align}
\vx_1^{\text{MAP}}
&=
\bigl(R + A^\top \Sigma^{-1} A\bigr)^{-1}
\bigl(R\hat{\vx}_1 + A^\top \Sigma^{-1}\vy\bigr)
\\
&=
\hat{\vx}_1
+
\bigl(R + A^\top \Sigma^{-1} A\bigr)^{-1}
A^\top \Sigma^{-1}
\bigl(\vy - A\hat{\vx}_1\bigr).
\end{align}
The second line makes explicit that the MAP solution is obtained by
adding a correction to $\hat{\vx}_1$. Using standard linear--Gaussian
identities, this correction term is
equivalent to the ProjFlow update
\begin{align}
\hat{\vx}_1^\star
=
\hat{\vx}_1
+
R^{-1}A^\top\bigl(AR^{-1}A^\top + \Sigma\bigr)^{-1}
\bigl(\vy - A\hat{\vx}_1\bigr),
\end{align}
showing that ProjFlow's projection step is exactly the MAP estimate of this linear--Gaussian model.

\section{Additional Method Details}
\label{sec:supp_methods}

\subsection{Formulating Teaser Applications as Linear Inverse Problems}
We briefly show how the additional teaser applications in Fig.~\ref{fig:teaser} fit into the unified linear model
$\vy = A\vx + \vepsilon$. Trajectory control and 2D-to-3D lifting are already described in the main paper. Here, we
detail the relative position constraint and looped motion.

\subsubsection{Relative Position Constraint}
We consider the case where the relative 3D position between two joints remains fixed, e.g., both wrists holding a rigid
object. Let $\vx_{n,j_a}, \vx_{n,j_b} \in \mathbb{R}^3$ denote the positions of joints $j_a$ and $j_b$ at frame $n$.
To keep their 3D offset fixed, we enforce for each frame
\begin{align}
\vx_{n,j_a} - \vx_{n,j_b} = \vd,
\end{align}
where $\vd = (d_x, d_y, d_z)^\top$ is the desired 3D offset vector. This is linear in the full motion vector $\vx$.
Stacking the constraints over all $N$ frames yields a standard linear inverse problem
\begin{align}
\vy_{\mathrm{rel}} = A_{\mathrm{rel}} \vx,
\end{align}
where $\vy_{\mathrm{rel}}$ is $\vd$ repeated $N$ times, so $\vy_{\mathrm{rel}} \in \mathbb{R}^{3N}$. The operator $A_{\mathrm{rel}} \in \mathbb{R}^{3N \times d}$ is a sparse matrix that, for each frame, subtracts the coordinates
of joint $j_b$ from those of joint $j_a$.

\subsubsection{Looped Motion}
To make a sequence loop seamlessly, we match the start and end poses. Let $\vx_0$ and $\vx_{N-1}$ be the first and last
frames of the motion, respectively. We impose the per-joint constraint
\begin{align}
\vx_0 - \vx_{N-1} = \mathbf{0},
\end{align}
which is again linear in $\vx$. Stacking these equations over all joints and spatial coordinates gives
\begin{align}
\mathbf{0} = A_{\mathrm{loop}} \vx,
\end{align}
where $A_{\mathrm{loop}} \in \mathbb{R}^{3J \times d}$ computes the difference between the first and last frames. In our framework, this loop-closure operator can simply be concatenated with other linear constraints by stacking its rows into the global observation matrix $A$.

\subsection{Detailed Formulation of Motion Inpainting}

\subsubsection{Pseudo-observations: linear interpolation and extrapolation}
We generate pseudo-observations by per-joint linear interpolation. For each joint, we scan all unobserved frames and, for a given unobserved frame, locate the nearest observed frame before it and the nearest observed frame after it. If both exist, the frame lies between two known points, and we define the pseudo-observation by linear interpolation between these two observations.

If the frame lies outside the observed range for that joint (before the first observation or after the last), interpolation is impossible. In this case, we perform extrapolation by copying the value of the single nearest observed frame. If a joint has no observations at all in the sequence, we leave it without pseudo-observations.

\subsubsection{Designing the adaptive variance}
\label{sec:pi2sigma}
Our inpainting strategy augments sparse hard keyframe constraints with “soft” pseudo-observations from interpolation. The key challenge is to modulate the influence of these soft guides: they should be trusted less (i) at frames with high motion curvature, where interpolation is unreliable, and (ii) late in sampling, when the model’s own prediction $\hat{\vx}_1$ is more reliable. We encode this behavior in a time-varying observation covariance $\Sigma^{(t)}$. Directly hand-designing variances $\sigma_i^2(t)$ is unintuitive, so we instead design a normalized trust score $\pi_i \in [0,1]$ and then derive the corresponding $\sigma_i^2(t)$.

To see the relation between $\pi_i$ and $\sigma_i^2(t)$, we first consider a simple Euclidean case. For motion inpainting, the observation operator is a diagonal mask matrix $A = M^{(t)}$. Assuming the Euclidean metric $R = I$, the ProjFlow update becomes
\begin{align}
 \hat{\vx}_1^\star
&= \hat{\vx}_1
 + M^{(t)}\bigl(M^{(t)} + \Sigma^{(t)}\bigr)^{-1}\bigl(\vy^{(t)} - A\hat{\vx}_1\bigr) \\
&= \Bigl(I - M^{(t)}\bigl(M^{(t)} + \Sigma^{(t)}\bigr)^{-1}M^{(t)}\Bigr)\hat{\vx}_1
\nonumber \\ 
&\quad+ M^{(t)}\bigl(M^{(t)} + \Sigma^{(t)}\bigr)^{-1}\vy^{(t)}.
\end{align}
In the inpainting setting, both $M^{(t)}$ and $\Sigma^{(t)}$ are diagonal, so this matrix equation decomposes into independent scalar updates. For an observed coordinate $i$ (i.e., $M^{(t)}_{ii}=1$) with $\Sigma^{(t)}_{ii} = \sigma_i^2(t)$, we obtain
\begin{align}
\hat{x}_{1,i}^\star
&= \left( 1 - \frac{1}{1 + \sigma_i^2(t)} \right) \hat{x}_{1,i}
 + \frac{1}{1 + \sigma_i^2(t)} y_i 
\end{align}
Thus, each updated coordinate is a weighted average of the model prediction $\hat{x}_{1,i}$ and the observation $y_i$. If we define the weight on the observation as
\begin{align}
\pi_{i,\text{Euclid}} \;\equiv\; \frac{1}{1 + \sigma_i^2(t)},
\end{align}
the update takes the intuitive form
\begin{align}
    \hat{x}_{1,i}^\star
    \;=\; (1-\pi_{i,\text{Euclid}})\,\hat{x}_{1,i}
        + \pi_{i,\text{Euclid}}\,y_i.
\end{align}
This shows that, in the Euclidean case, the “weight on data” for an active coordinate is exactly $\pi_{i,\text{Euclid}} = 1/(1+\sigma_i^2(t))$.

We now extend this idea to the kinematics-aware metric. The ProjFlow update becomes
{\small
\begin{align}
\hat{\vx}_1^\star
&= \hat{\vx}_1
 + R^{-1}{M^{(t)}}^\top\!\Bigl(M^{(t)}R^{-1}{M^{(t)}}^\top + \Sigma^{(t)}\Bigr)^{-1}
 \bigl(\vy^{(t)}-M^{(t)}\hat{\vx}_1\bigr) \\
&= \Bigl(
I - R^{-1}{M^{(t)}}^\top\!\Bigl(M^{(t)}R^{-1}{M^{(t)}}^\top+\Sigma^{(t)}\Bigr)^{-1} M^{(t)}
\Bigr)\hat{\vx}_1 \nonumber \\
&\quad
 + R^{-1}{M^{(t)}}^\top\!\Bigl(M^{(t)}R^{-1}{M^{(t)}}^\top+\Sigma^{(t)}\Bigr)^{-1}\vy^{(t)}.
\label{eq:kkt_update_selector}
\end{align}
}
Here, $R^{-1}$ is dense along joint dimensions, so corrections propagate across joints, while we still choose $\Sigma^{(t)}$ to be diagonal, with each coordinate (frame–joint–axis) having its own variance. We therefore design a dimensionless trust score $\pi_i\in[0,1]$ for each active row $i$, and convert it into a variance that is consistent with the metric $R$. 

Let $r_i:= [\mathrm{diag}(R^{-1})]_i > 0$. If only row $i$ were active (i.e., $M^{(t)} = \ve_i^\top$), the measurement-space gain of the ProjFlow update is
\begin{align}
\pi_i
\;=\; \frac{r_i}{\,r_i + \sigma_i^2(t)\,}.
\end{align}
Solving for $\sigma_i^2(t)$ yields
\begin{equation}
\Sigma^{(t)}_{ii} \;=\; \sigma_i^2(t)
 \;=\; r_i\!\left(\frac{1}{\pi_i}-1\right).
\label{eq:pi2sigma_general}
\end{equation}
Note that when $R = I$, we have $r_i = 1$, and \eqref{eq:pi2sigma_general} reduces to
$\pi_i = 1/(1+\sigma_i^2(t))$, matching the Euclidean case.

\subsubsection{Computing the variance from the trust score for multiple joints}
To obtain the per-element trust scores $\pi_i$, we first compute a frame-level base trust
\begin{align}
\tilde{\pi}_n^{(t)}
\;=\;
\tau(t)\,\frac{c_0}{1+\lambda_s\,(s_n(\hat{\vx}_1)/s_{\text{med}})^p},
\end{align}
where $n$ indexes frames, $s_n(\hat{\vx}_1)$ is the curvature at frame $n$, $s_{\text{med}}$ is the median curvature over the sequence, and $c_0,\lambda_s,p$ are hyperparameters. This $\tilde{\pi}_n^{(t)}$ is the total “trust budget” for all active pseudo-observations in frame $n$. If only one joint has an active pseudo-observation at that frame, we simply set $\pi_i = \tilde{\pi}_n^{(t)}$.

If multiple joints are active in frame $n$, we distribute the frame-level budget across them according to their influence in the kinematics-aware metric $R$. Intuitively, we want to assign less trust to high-influence joints (e.g., pelvis) and more trust to low-influence joints (e.g., wrists). Let $\mathcal{H}_n$ be the set of joints $j$ with an active pseudo-observation in frame $n$, and $m_n = |\mathcal{H}_n|$. Recall that
\begin{align}
R = w_{\text{kin}}(I_{3}\otimes I_{N}\otimes L_{\text{kin}})+\lambda I_{d},
\end{align}
and define the joint-only component
\begin{align}
R_J = w_{\text{kin}}L_{\text{kin}} + \lambda I_J.
\end{align}
From $R_J$, we define a per-joint weight as
\begin{align}
q_j \;:=\; \frac{1}{\big\lVert \mathbf{c}_j \big\rVert_2},
\end{align}
where $\mathbf{c}_j$ denotes the $j$-th column of $R_J^{-1}$. In other words, $q_j$ is the reciprocal of the Euclidean norm of the $j$-th column of $R_J^{-1}$. Joints with large global influence yield columns with large norms and therefore smaller $q_j$,  whereas low‑influence joints yield smaller column norms and thus larger $q_j$.

We then distribute the frame budget proportionally to these weights. For an element $i$ corresponding to joint $j \in \mathcal{H}_n$, we set
\begin{align}
\pi_i
\;=\;
\mathrm{clip}\!\left(
\tilde{\pi}_n^{(t)}\,
\frac{q_j}{\sum_{k\in\mathcal{H}_n} q_k},
\;\pi_{\min},\;\pi_{\max}
\right).
\end{align}
Ignoring clipping, this construction preserves the frame-level budget,
$\sum_{i \in \mathcal{H}_n}\pi_i = \tilde{\pi}_n^{(t)}$, while assigning lower trust to high-influence joints and higher trust to low-influence ones. Finally, these $\pi_i$ are converted to variances $\sigma_i^2(t)$ via ~\eqref{eq:pi2sigma_general}, yielding the diagonal entries of $\Sigma^{(t)}$ for the active pseudo-observations.

\section{Implementation Details}
\label{sec:supp_impl}

\subsection{Application I: Motion Inpainting via Masked Pseudo-observations}
The hyperparameters used for motion inpainting are summarized in Table~\ref{tab:inpaint_hparams}. We use the same values for all inpainting experiments, including the main comparison in Table~\ref{tab:control_noAITS} and the ablation study in Table~\ref{tab:ablation_control_67}. We set the number of ODE sampling steps to $T = 100$, which corresponds to 100 function evaluations. The kinematics-aware metric $R$ is parameterized with $w_{\text{kin}} = 10.0$ and $\lambda = 1.0$. The dynamic masking radius shrinks linearly from $l_{\max} = 10$ to $l_{\min} = 3$ frames over time. For recomposition, we adopt the stochastic step from FlowDPS~\cite{kim2025flowdps} with the noise-mixing schedule $\eta_t = 1 - \sigma_{t+\Delta t}$.

\begin{table}[ht]
\centering
\small
\resizebox{\linewidth}{!}{
\begin{tabular}{lllc}
\toprule
\textbf{Block} & \textbf{Name} & \textbf{Symbol} & \textbf{Value} \\
\midrule
\multirowcell{2}{\textit{Kinematics-aware} \\ \textit{metric}}
& joint coupling weight & $w_{\text{kin}}$& 10.0 \\
& ridge  & $\lambda$& 1.0 \\
\midrule
\multirowcell{2}{\textit{Dynamic Masking}}
& min radius (frames) & $l_{\min}$& 3 \\
& max radius (frames) & $l_{\max}$& 10 \\
\midrule
\multirowcell{5}{\textit{Adaptive Variance}}
& time base & $\tau_{\min}$& 0.1 \\
& strength & $c_0$& 3.0 \\
& curvature gain & $\lambda_s$& 1.0 \\
& curvature power & $p$& 2.0 \\
& trust clipping & $[\pi_{\min},\pi_{\max}]$ & [0.02, 1.0] \\
\midrule
\multirowcell{2}{\textit{ODE sampling}}
& NFE  & $T$ & 100 \\
& noise mixing & $\eta_t$ & $\eta_t=1 - \sigma_{t+\Delta t}$ \\
\bottomrule
\end{tabular}}
\caption{\textbf{Hyperparameters for motion inpainting.}}
\label{tab:inpaint_hparams}
\end{table}

\subsection{Application II: 2D-to-3D Lifting via Linear Projection Measurements}
For the 2D-to-3D motion lifting application, we reuse the ODE sampler hyperparameters from the inpainting task. We again set $T = 100$ sampling steps and use the FlowDPS~\citep{kim2025flowdps} noise-mixing schedule $\eta_t = 1 - \sigma_{t+\Delta t}$. The kinematics-aware metric $R$ also uses the same values $w_{\text{kin}} = 10.0$ and $\lambda = 1.0$ as in the inpainting experiments, without additional tuning for this task.

\section{Additional Quantitative Results}
\label{sec:supp_results}

\subsection{Inference Speed Comparison}
\begin{figure}[t]
\centering
\includegraphics[width=\linewidth]{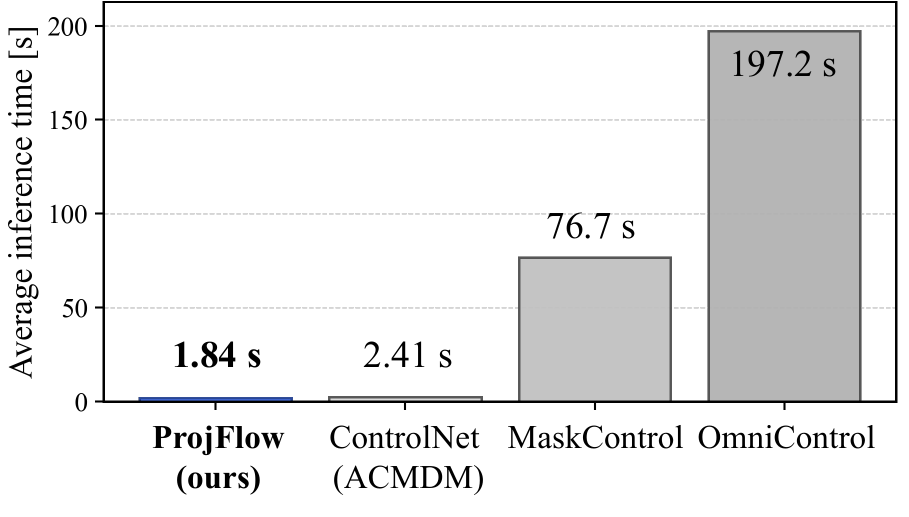}
\vspace{-0.75em}
\caption{\textbf{Average inference time per 196-frame sample .}}
\label{fig:inference_time}
\end{figure}

We compare the inference speed of ProjFlow against training-based controllers that use the same backbone. Figure~\ref{fig:inference_time} reports the average wall-clock time required to generate one 196-frame motion sample on a single A100 GPU. The x-axis labels in the figure use abbreviated names: ProjFlow (ours) and ControlNet (ACMDM) correspond to ACMDM\text{-}S\text{-}PS22+ProjFlow and ACMDM\text{-}S\text{-}PS22+ControlNet, respectively. We use the original settings from each paper whenever they are specified. For ControlNet~\cite{meng2025acmdm}, whose sampling schedule is not detailed, we match our ProjFlow configuration for fairness: both ProjFlow and ControlNet use 100 Euler steps. OmniControl~\cite{xie2024omnicontrol} is evaluated with 1{,}000 sampling steps. MaskControl~\cite{pinyoanuntapong2024controlmm} uses 10 sampling steps, with 100 logits-optimization steps at each unmasking step and 600 optimization steps at the final unmasking step as described in the original paper. 

Under these settings, ProjFlow achieves an average inference time of \textbf{1.84\,s} per sample and is the fastest among all compared methods. Notably, even though ProjFlow and ControlNet (ACMDM) share the same 100-step sampling schedule, ProjFlow runs faster because it keeps the original backbone unchanged, whereas ControlNet attaches an additional conditioning branch that increases model size and inference cost.

\begin{figure*}[t]
\centering
\includegraphics[width=\linewidth]{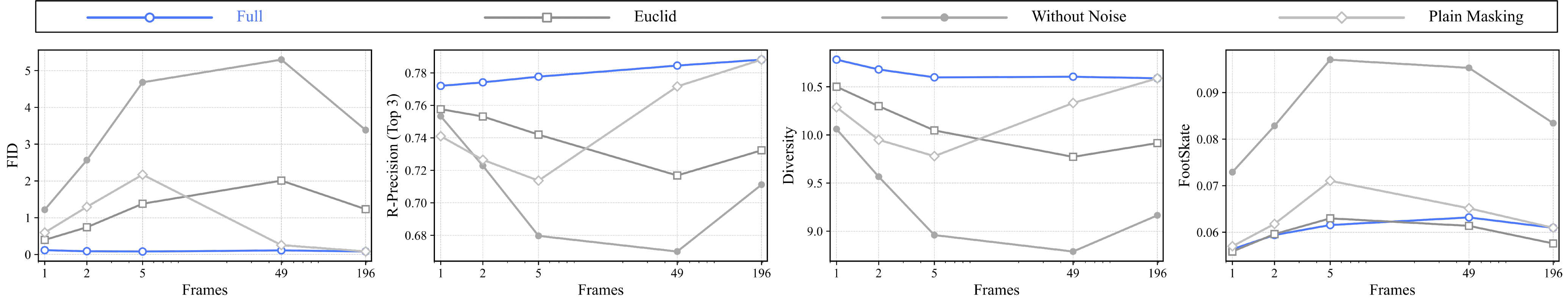}
\caption{\textbf{Ablation of ProjFlow components vs.\ control intensity on motion inpainting.}
We compare the full model (\texttt{Full}) against variants that
(i) replace the kinematics-aware metric with a Euclidean metric (\texttt{Euclid}),
(ii) remove the stochastic noise-mixing step (\texttt{Without Noise}), or
(iii) disable pseudo-observations and rely only on hard keyframes (\texttt{Plain Masking}).}
\label{fig:ablation_intensity_performance}
\end{figure*}

\subsection{Detailed Results of Ablation Study}
Figure~\ref{fig:ablation_intensity_performance} summarizes how each ProjFlow component
affects robustness to control intensity on the motion inpainting task.
The model \texttt{Full} is compared with three ablations:
\texttt{Euclid}, which uses a standard Euclidean metric instead of the kinematics-aware metric;
\texttt{Without Noise}, which removes the stochastic noise-mixing step; and
\texttt{Plain Masking}, which removes pseudo-observations and uses only hard keyframes.
When observations are sparse, both the Euclidean metric and the deterministic recomposition
(\texttt{Without Noise}) noticeably degrade realism, and the \texttt{Plain Masking} variant
performs worst, confirming the importance of our pseudo-observations.
Table~\ref{tab:supp_control_ablation} reports the full per-joint numbers:
our \texttt{Full} model consistently attains the best FID and R-Precision across all controlled joints,
while keeping trajectory, location, and average control errors at zero.

\input{table/supp_ablation}

\subsection{Detailed Results of Motion Inpainting}
In the main paper Table~\ref{tab:control_noAITS}, we presented a summrized version of the controllable motion generation
results. Table~\ref{tab:supp_control_full} provides the complete per-joint evaluation,
following the OmniControl~\citep{xie2024omnicontrol} protocol. Across all controlled joints,
ProjFlow achieves zero trajectory, location, and average errors, while its FID, R-Precision,
and diversity scores remain in the same band as the strongest training-based controllers.
This shows that enforcing exact spatial constraints with ProjFlow does not come at the
expense of motion realism.

\input{table/supp_control}

\subsection{Evaluation on Legacy Metrics}
\citet{meng2025rethinking} recently highlighted several shortcomings in the conventional
HumanML3D evaluation protocol and proposed revised metrics, which we adopt for the
main results in the paper. However, many prior works~\citep{tevet2023human,xie2024omnicontrol,pinyoanuntapong2024mmm, dai2024motionlcm, pinyoanuntapong2024controlmm, shafir2024priormdm, karunratanakul2023gmd} still report performance using
the legacy protocol, making direct comparison otherwise impossible. To broaden the set
of comparable baselines, we therefore also evaluate ProjFlow and existing methods under
the original evaluation setup. The results are summarized in Table~\ref{tab:control_noAITS263}.
Under this legacy protocol, ProjFlow remains competitive with strong training-based
controllers while retaining its zero-shot nature and exact constraint satisfaction.

\input{table/control_263dim}

\section{Additional qualitative results.}
\label{sec:supp_results_video}
The supplementary material includes a browsable demo page that collects our qualitative videos (open \texttt{index.html} in a web browser). This page organizes examples by task: the four control scenarios from Fig.\ref{fig:teaser}, trajectory-control benchmarks comparing ProjFlow with OmniControl\citep{xie2024omnicontrol} and MaskControl\citep{xie2024omnicontrol}, and 2D-to-3D lifting comparisons against Sketch2Anim\citep{zhong2025sketch2anim}. We refer readers to this page for a more complete visual impression of ProjFlow’s behavior.

%% file: table/supp_ablation.tex
\begin{table*}[t]
    \centering
    \caption{\textbf{Ablation of ACMDM-S-PS22+ProjFlow on HumanML3D.} Methods are evaluated on all joints and reported per controlled joint. \textbf{bold} face / \underline{underline} indicates the best/2\textsuperscript{nd} results if applied.}
    \renewcommand{\arraystretch}{1}
    \resizebox{1\linewidth}{!}{
    \begin{tabular}{clccccccc}
    \toprule
       \multirowcell{2}{Controlling\\ Joint} & \multirow{2}{*}{Methods} & \multirow{2}{*}{FID$\downarrow$}& R-Precision & \multirow{2}{*}{Diversity$\rightarrow$} & Foot Skating& \multirow{2}{*}{Traj. err.$\downarrow$} & \multirow{2}{*}{Loc. err.$\downarrow$} & \multirow{2}{*}{Avg. err.$\downarrow$}\\
    ~ & && Top 3 & & Ratio.$\downarrow$ & & & \\
    \midrule
    &GT &$0.000$ &$0.795$ &$10.455$ &-&$0.0000$ &$0.0000$ &$0.0000$\\
    \midrule
    \multirow{4}{*}{\textbf{Pelvis}}
      &\textbf{ACMDM\text{-}S\text{-}PS22+ProjFlow (Full)} &\textbf{0.107} &\textbf{0.784} &\textbf{10.645} &\underline{0.0630} &0.0000 &0.0000 &0.0000\\
      &ACMDM\text{-}S\text{-}PS22+ProjFlow (Euclid)        &4.360 &0.686 &8.953  &\textbf{0.0550} &0.0000 &0.0000 &0.0000\\
      &ACMDM\text{-}S\text{-}PS22+ProjFlow (w/o noise)     &2.439 &\underline{0.734} &9.666  &0.0960 &0.0000 &0.0000 &0.0000\\
      &ACMDM\text{-}S\text{-}PS22+ProjFlow (Plain Masking) &\underline{2.091} &0.726 &\underline{9.838}  &0.0658 &0.0000 &0.0000 &0.0000\\
    \midrule

    \multirow{4}{*}{\textbf{Left foot}}
      &\textbf{ACMDM\text{-}S\text{-}PS22+ProjFlow (Full)} &\textbf{0.095} &\textbf{0.771} &\textbf{10.644} &\textbf{0.0609} &0.0000 &0.0000 &0.0000\\
      &ACMDM\text{-}S\text{-}PS22+ProjFlow (Euclid)        &\underline{0.476} &0.743 &\underline{10.399} &\underline{0.0643} &0.0000 &0.0000 &0.0000\\
      &ACMDM\text{-}S\text{-}PS22+ProjFlow (w/o noise)     &4.450 &0.680 &9.209  &0.0969 &0.0000 &0.0000 &0.0000\\
      &ACMDM\text{-}S\text{-}PS22+ProjFlow (Plain Masking) &0.576 &\underline{0.746} &10.309 &0.0681 &0.0000 &0.0000 &0.0000\\
    \midrule

    \multirow{4}{*}{\textbf{Right foot}}
      &\textbf{ACMDM\text{-}S\text{-}PS22+ProjFlow (Full)} &\textbf{0.096} &\textbf{0.770} &\textbf{10.651} &\textbf{0.0613} &0.0000 &0.0000 &0.0000\\
      &ACMDM\text{-}S\text{-}PS22+ProjFlow (Euclid)        &\underline{0.486} &0.745 &\underline{10.359} &\underline{0.0655} &0.0000 &0.0000 &0.0000\\
      &ACMDM\text{-}S\text{-}PS22+ProjFlow (w/o noise)     &4.805 &0.673 &9.129  &0.0944 &0.0000 &0.0000 &0.0000\\
      &ACMDM\text{-}S\text{-}PS22+ProjFlow (Plain Masking) &0.520 &\underline{0.748} &10.335 &0.0675 &0.0000 &0.0000 &0.0000\\
    \midrule

    \multirow{4}{*}{\textbf{Head}}
      &\textbf{ACMDM\text{-}S\text{-}PS22+ProjFlow (Full)} &\textbf{0.099} &\textbf{0.788} &\textbf{10.754} &0.0595 &0.0000 &0.0000 &0.0000\\
      &ACMDM\text{-}S\text{-}PS22+ProjFlow (Euclid)        &\underline{0.560} &\underline{0.761} &\underline{10.332} &\textbf{0.0547} &0.0000 &0.0000 &0.0000\\
      &ACMDM\text{-}S\text{-}PS22+ProjFlow (w/o noise)     &1.852 &0.750 &9.714  &0.0706 &0.0000 &0.0000 &0.0000\\
      &ACMDM\text{-}S\text{-}PS22+ProjFlow (Plain Masking) &1.076 &0.751 &10.175 &\underline{0.0594} &0.0000 &0.0000 &0.0000\\
    \midrule

    \multirow{4}{*}{\textbf{Left wrist}}
      &\textbf{ACMDM\text{-}S\text{-}PS22+ProjFlow (Full)} &\textbf{0.089} &\textbf{0.783} &\textbf{10.601} &\underline{0.0586} &0.0000 &0.0000 &0.0000\\
      &ACMDM\text{-}S\text{-}PS22+ProjFlow (Euclid)        &0.524 &0.754 &\underline{10.256} &\textbf{0.0583} &0.0000 &0.0000 &0.0000\\
      &ACMDM\text{-}S\text{-}PS22+ProjFlow (w/o noise)     &3.507 &0.703 &9.019  &0.0801 &0.0000 &0.0000 &0.0000\\
      &ACMDM\text{-}S\text{-}PS22+ProjFlow (Plain Masking) &\underline{0.506} &\underline{0.759} &10.242 &0.0590 &0.0000 &0.0000 &0.0000\\
    \midrule

    \multirow{4}{*}{\textbf{Right wrist}}
      &\textbf{ACMDM\text{-}S\text{-}PS22+ProjFlow (Full)} &\textbf{0.096} &\textbf{0.780} &\textbf{10.610} &\textbf{0.0584} &0.0000 &0.0000 &0.0000\\
      &ACMDM\text{-}S\text{-}PS22+ProjFlow (Euclid)        &\underline{0.506} &0.753 &\underline{10.343} &\underline{0.0591} &0.0000 &0.0000 &0.0000\\
      &ACMDM\text{-}S\text{-}PS22+ProjFlow (w/o noise)     &3.522 &0.705 &9.111  &0.0799 &0.0000 &0.0000 &0.0000\\
      &ACMDM\text{-}S\text{-}PS22+ProjFlow (Plain Masking) &0.514 &\underline{0.759} &10.224 &0.0594 &0.0000 &0.0000 &0.0000\\
    \midrule

    \multirow{4}{*}{\textbf{Average}}
      &\textbf{ACMDM\text{-}S\text{-}PS22+ProjFlow (Full)} &\textbf{0.097} &\textbf{0.779} &\textbf{10.651} &\underline{0.0603} &0.0000 &0.0000 &0.0000\\
      &ACMDM\text{-}S\text{-}PS22+ProjFlow (Euclid)        &1.152 &0.740 &10.107 &\textbf{0.0595} &0.0000 &0.0000 &0.0000\\
      &ACMDM\text{-}S\text{-}PS22+ProjFlow (w/o noise)     &3.429 &0.707 &9.308  &0.0863 &0.0000 &0.0000 &0.0000\\
      &ACMDM\text{-}S\text{-}PS22+ProjFlow (Plain Masking) &\underline{0.881} &\underline{0.748} &\underline{10.187} &0.0632 &0.0000 &0.0000 &0.0000\\
    \bottomrule
    \end{tabular}}
    \label{tab:supp_control_ablation}
\end{table*}

%% file: table/supp_control.tex
\begin{table*}[t]
    \centering
    \caption{\textbf{Quantitative text-conditioned motion generation with spatial control signals and upper-body editing on HumanML3D.} In the first section, methods are trained and evaluated solely on pelvis controls. In the middle section, methods are trained on all joints and evaluated separately on each controlled joint. \textbf{bold} face / \underline{underline} indicates the best/2\textsuperscript{nd} results.}
    \renewcommand{\arraystretch}{1.0}
    \resizebox{1\linewidth}{!}{
    \begin{tabular}{clcccccccc}
    \toprule
       \multirowcell{2}{Controlling\\ Joint} & \multirow{2}{*}{Methods} & \multirow{2}{*}{Zero-shot?}& \multirow{2}{*}{FID$\downarrow$}& R-Precision & \multirow{2}{*}{Diversity$\rightarrow$} & Foot Skating& \multirow{2}{*}{Traj. err.$\downarrow$} & \multirow{2}{*}{Loc. err.$\downarrow$} & \multirow{2}{*}{Avg. err.$\downarrow$}\\
    ~ & &&& Top 3 & & Ratio.$\downarrow$ & & & \\
    \midrule
    &GT &$-$ &$0.000$ &$0.795$ &$10.455$ &-&$0.000$ &$0.000$ &$0.000$\\
    \midrule
    \multirow{7}{*}{\parbox{2.2cm}{\centering \textbf{Train\\On\\Pelvis}}}
    &MDM~\citep{tevet2023human} &\ding{51} &$1.792$ &$0.673$ &$9.131$ &$0.1019$ &$0.4022$ &$0.3076$ &$0.5959$\\
    &PriorMDM~\citep{shafir2024priormdm} &\ding{55} &$0.393$ &$0.707$ &$9.847$ &$0.0897$&$0.3457$ &$0.2132$ &$0.4417$\\
    &GMD~\citep{karunratanakul2023gmd}  &\ding{51} &$0.238$ &$0.763$ &$10.011$ &$0.1009$&$0.0931$ &$0.0321$ &$0.1439$\\
    &OmniContol~\citep{xie2024omnicontrol} &\ding{55} &$0.081$ &$0.789$ &$10.323$  &$\underline{0.0547}$ &$0.0387$ &$0.0096$ &$0.0338$\\
    &MotionLCM V2+CtrlNet~\citep{dai2024motionlcm} &\ding{55} &$3.978$ &$0.738$ &$9.249$ &$0.0901$ &$0.1080$ &$0.0581$&$0.1386$ \\
    &MaskControl~\citep{pinyoanuntapong2024controlmm} &\ding{55}&$\mathbf{0.066}$ &$0.799$ &$10.474$ &$\mathbf{0.0543}$ &$\mathbf{0.0000}$ &$\mathbf{0.0000}$& $0.0093$\\
    &ACMDM-S-PS22+CtrlNet~\citep{meng2025acmdm} &\ding{55} &$\underline{0.067}$ &$\mathbf{0.805}$ &$\underline{10.481}$ &$0.0591$ &$0.0075$ &$0.0010$ &$0.0100$\\
    &ACMDM-S-PS22+DNO~\citep{karunratanakul2024dno} &\ding{51} &$0.151$ &$\underline{0.802}$ &$-$ &$0.0610$ &$\underline{0.0027}$ &$\underline{0.0002}$ &$\underline{0.0089}$\\
    &\textbf{ACMDM-S-PS22+ProjFlow (ours)} &\ding{51} &$0.107$ &$0.784$ &$\mathbf{10.645}$ &$0.0630$ &$\mathbf{0.0000}$ &$\mathbf{0.0000}$ &$\mathbf{0.0000}$\\
    \midrule
    \midrule

    \multirow{6}{*}{\textbf{Pelvis}}
    &OmniContol~\citep{xie2024omnicontrol} &\ding{55} &$0.135$ &$0.790$ &$10.314$ &$\underline{0.0571}$ &$0.0404$ &$0.0085$ &$0.0367$\\
    &MotionLCM V2+CtrlNet~\citep{dai2024motionlcm} &\ding{55}&$4.726$&$0.713$&$9.209$ &$0.1162$&$0.1617$ &$0.0841$ &$0.1838$\\
    &MaskControl~\citep{pinyoanuntapong2024controlmm} &\ding{55} &$\underline{0.087}$ &$0.795$ &$10.168$ &$\mathbf{0.0544}$ &$\underline{0.0003}$ &$\mathbf{0.0000}$ &$0.0114$\\
    &ACMDM-S-PS22+CtrlNet~\citep{meng2025acmdm} &\ding{55} &$\mathbf{0.075}$ &$\mathbf{0.805}$ &$\underline{10.536}$ &$0.0603$ &$0.0081$ &$0.0011$ &$0.0134$\\
    &ACMDM-S-PS22+DNO~\citep{karunratanakul2024dno} &\ding{51} &$0.151$ &$\underline{0.802}$ &$-$ &$0.0610$ &$\underline{0.0027}$ &$\underline{0.0002}$ &$\underline{0.0089}$\\
    &\textbf{ACMDM-S-PS22+ProjFlow (ours)} &\ding{51} &$0.107$ &$0.784$ &$\mathbf{10.645}$ &$0.0630$ &$\mathbf{0.0000}$ &$\mathbf{0.0000}$ &$\mathbf{0.0000}$\\
    \midrule

    \multirow{6}{*}{\textbf{Left foot}}
    &OmniContol~\citep{xie2024omnicontrol} &\ding{55} &$0.093$ &$0.794$ &$10.338$ &$0.0692$&$0.0594$ &$0.0094$ &$0.0314$\\
    &MotionLCM V2+CtrlNet~\citep{dai2024motionlcm} &\ding{55}&$4.810$ &$0.706$ &$9.158$ &$0.1047$&$0.2607$ &$0.1229$ &$0.2304$\\
    &MaskControl~\citep{pinyoanuntapong2024controlmm} &\ding{55} &$\underline{0.074}$ &$0.793$ &$10.241$ &$\mathbf{0.0561}$ &$\mathbf{0.0000}$ &$\mathbf{0.0000}$ &$\underline{0.0066}$\\
    &ACMDM-S-PS22+CtrlNet~\citep{meng2025acmdm} &\ding{55} &$\mathbf{0.063}$ &$\mathbf{0.800}$ &$\underline{10.542}$ &$\underline{0.0590}$ &$0.0186$ &$0.0034$ &$0.0240$\\
    &ACMDM-S-PS22+DNO~\citep{karunratanakul2024dno} &\ding{51} &$0.147$ &$\underline{0.799}$ &$-$ &$0.0602$ &$\underline{0.0082}$ &$\underline{0.0003}$ &$0.0133$\\
    &\textbf{ACMDM-S-PS22+ProjFlow (ours)} &\ding{51} &$0.095$ &$0.771$ &$\mathbf{10.644}$ &$0.0609$ &$\mathbf{0.0000}$ &$\mathbf{0.0000}$ &$\mathbf{0.0000}$\\
    \midrule

    \multirow{6}{*}{\textbf{Right foot}}
    &OmniContol~\citep{xie2024omnicontrol} &\ding{55} &$0.137$ &$0.798$ &$10.241$ &$0.0668$&$0.0666$ &$0.0120$ &$0.0334$\\
    &MotionLCM V2+CtrlNet~\citep{dai2024motionlcm} &\ding{55} &$4.756$ &$0.705$ &$9.303$ &$0.1026$&$0.2459$ &$0.1127$ &$0.2278$\\
    &MaskControl~\citep{pinyoanuntapong2024controlmm} &\ding{55} &$\underline{0.080}$ &$0.793$ &$10.159$ &$\mathbf{0.0552}$ &$\mathbf{0.0000}$ &$\mathbf{0.0000}$ &$\underline{0.0062}$\\
    &ACMDM-S-PS22+CtrlNet~\citep{meng2025acmdm} &\ding{55} &$\mathbf{0.071}$ &$\mathbf{0.803}$ &$\underline{10.591}$ &$\underline{0.0583}$ &$0.0205$ &$0.0030$ &$0.0251$\\
    &ACMDM-S-PS22+DNO~\citep{karunratanakul2024dno} &\ding{51} &$0.153$ &$\underline{0.800}$ &$-$ &$0.0597$ &$\underline{0.0086}$ &$\underline{0.0003}$ &$0.0138$\\
    &\textbf{ACMDM-S-PS22+ProjFlow (ours)} &\ding{51} &$0.096$ &$0.770$ &$\mathbf{10.651}$ &$0.0613$ &$\mathbf{0.0000}$ &$\mathbf{0.0000}$ &$\mathbf{0.0000}$\\

    \midrule

    \multirow{6}{*}{\textbf{Head}}
    &OmniContol~\citep{xie2024omnicontrol} &\ding{55} &$0.146$ &$0.796$ &$10.239$ &$\underline{0.0556}$ &$0.0422$ &$0.0079$ &$0.0349$\\
    &MotionLCM V2+CtrlNet~\citep{dai2024motionlcm} &\ding{55} &$4.580$ &$0.715$ &$9.278$ &$0.1138$&$0.1971$ &$0.0977$ &$0.2136$\\
    &MaskControl~\citep{pinyoanuntapong2024controlmm} &\ding{55} &$\underline{0.090}$ &$0.797$ &$10.131$ &$\mathbf{0.0531}$ &$\mathbf{0.0000}$ &$\mathbf{0.0000}$ &$\underline{0.0064}$\\
    &ACMDM-S-PS22+CtrlNet~\citep{meng2025acmdm} &\ding{55} &$\mathbf{0.081}$ &$\mathbf{0.805}$ &$\underline{10.520}$ &$0.0598$ &$0.0051$ &$0.0009$ &$0.0152$\\
    &ACMDM-S-PS22+DNO~\citep{karunratanakul2024dno} &\ding{51} &$0.138$ &$\underline{0.801}$ &$-$ &$0.0591$ &$\underline{0.0025}$ &$\underline{0.0002}$ &$0.0084$\\
    &\textbf{ACMDM-S-PS22+ProjFlow (ours)} &\ding{51} &$0.099$ &$0.788$ &$\mathbf{10.754}$ &$0.0595$ &$\mathbf{0.0000}$ &$\mathbf{0.0000}$ &$\mathbf{0.0000}$\\
    \midrule

    \multirow{6}{*}{\textbf{Left wrist}}
    &OmniContol~\citep{xie2024omnicontrol} &\ding{55} &$0.119$ &$0.783$ &$10.217$ & $\underline{0.0562}$&$0.0801$ &$0.0134$ &$0.0529$\\
    &MotionLCM V2+CtrlNet~\citep{dai2024motionlcm} &\ding{55} &$4.103$ &$0.726$ &$9.188$ &$0.1167$ &$0.3965$ &$0.1912$ &$0.3150$\\
    &MaskControl~\citep{pinyoanuntapong2024controlmm} &\ding{55} &$0.118$ &$0.797$ &$10.153$ &$\mathbf{0.0546}$ &$\mathbf{0.0000}$ &$\mathbf{0.0000}$ &$\underline{0.0044}$\\
    &ACMDM-S-PS22+CtrlNet~\citep{meng2025acmdm} &\ding{55} &$\mathbf{0.065}$ &$\mathbf{0.804}$ &$\underline{10.480}$ &$0.0604$ &$0.0085$ &$0.0014$ &$0.0206$\\
    &ACMDM-S-PS22+DNO~\citep{karunratanakul2024dno} &\ding{51} &$0.149$ &$\underline{0.799}$ &$-$ &$0.0600$ &$\underline{0.0076}$ &$\underline{0.0004}$ &$0.0138$\\
    &\textbf{ACMDM-S-PS22+ProjFlow (ours)} &\ding{51} &$\underline{0.089}$ &$0.783$ &$\mathbf{10.601}$ &$0.0586$ &$\mathbf{0.0000}$ &$\mathbf{0.0000}$ &$\mathbf{0.0000}$\\
    \midrule
    
    \multirow{6}{*}{\textbf{Right wrist}}
    &OmniContol~\citep{xie2024omnicontrol} &\ding{55} &$0.128$ &$0.792$ &$10.309$ & $0.0601$&$ 0.0813$ &$0.0127$ &$0.0519$\\
    &MotionLCM V2+CtrlNet~\citep{dai2024motionlcm} &\ding{55} &$4.051$ &$0.725$ &$9.242$ &$0.1176$ &$ 0.3822$ &$0.1806$ &$0.3079$\\
    &MaskControl~\citep{pinyoanuntapong2024controlmm} &\ding{55} &$0.121$ &$0.797$ &$10.105$ &$\mathbf{0.0537}$ &$\mathbf{0.0000}$ &$\mathbf{0.0000}$ &$\underline{0.0044}$\\
    &ACMDM-S-PS22+CtrlNet~\citep{meng2025acmdm} &\ding{55} &$\mathbf{0.066}$ &$\mathbf{0.802}$ &$\underline{10.484}$ &$0.0599$ &$0.0091$ &$0.0016$ &$0.0201$\\
    &ACMDM-S-PS22+DNO~\citep{karunratanakul2024dno} &\ding{51} &$0.143$ &$\underline{0.798}$ &$-$ &$0.0598$ &$\underline{0.0081}$ &$\underline{0.0004}$ &$0.0142$\\
    &\textbf{ACMDM-S-PS22+ProjFlow (ours)} &\ding{51} &$\underline{0.096}$ &$0.780$ &$\mathbf{10.610}$ &$\underline{0.0584}$ &$\mathbf{0.0000}$ &$\mathbf{0.0000}$ &$\mathbf{0.0000}$\\

    \midrule

    \multirow{6}{*}{\textbf{Average}}
    &OmniContol~\citep{xie2024omnicontrol} &\ding{55} &$0.126$ &$0.792$ &$10.276$ & $0.0608$&$0.0617$ &$0.0107$ &$0.0404$\\
    &MotionLCM V2+CtrlNet~\citep{dai2024motionlcm} &\ding{55} &$4.504$ &$0.715$ &$9.230$ &0.1119 &$0.2740$ &$0.1315$ &$0.2464$\\
    &MaskControl~\citep{pinyoanuntapong2024controlmm} &\ding{55} &$\underline{0.095}$ &$0.795$ &$10.159$ &$\mathbf{0.0545}$ &$\underline{0.0001}$ &$\mathbf{0.0000}$ &$\underline{0.0065}$\\
    &ACMDM-S-PS22+CtrlNet~\citep{meng2025acmdm} &\ding{55} &$\mathbf{0.070}$ &$\mathbf{0.803}$ &$\underline{10.526}$ &$\underline{0.0596}$ &$0.0117$ &$0.0019$ &$0.0197$\\
    &ACMDM-S-PS22+DNO~\citep{karunratanakul2024dno} &\ding{51} &$0.147$ &$\underline{0.800}$ &$-$ &$0.0600$ &$0.0034$ &$\underline{0.0003}$ &$0.0121$\\
    &\textbf{ACMDM-S-PS22+ProjFlow (ours)} &\ding{51} &$0.097$ &$0.779$ &$\mathbf{10.651}$ &$0.0603$ &$\mathbf{0.0000}$ &$\mathbf{0.0000}$ &$\mathbf{0.0000}$\\
    \bottomrule
    \end{tabular}}
    \label{tab:supp_control_full}
\end{table*}

%% file: table/control_263dim.tex
\begin{table*}[t]
    \centering
    \caption{\textbf{Quantitative text-conditioned motion generation with spatial control signals and upper-body editing on HumanML3D~\cite{guo2022humanml3d}.} The first section covers pelvis-only control; the middle section shows the average for all joints. The last section presents upper-body editing results. 
    \textbf{bold} face / \underline{underline} indicates the best/2\textsuperscript{nd} results.}
    \renewcommand{\arraystretch}{1}
    \resizebox{1\linewidth}{!}{
    \begin{tabular}{clcccccccc}
    \toprule
       \multirowcell{2}{Controlling\\ Joint} & \multirow{2}{*}{Methods} & \multirow{2}{*}{Zero-shot?}& \multirow{2}{*}{FID$\downarrow$}& R-Precision & \multirow{2}{*}{Diversity$\rightarrow$} & Foot Skating& \multirow{2}{*}{Traj. err.$\downarrow$} & \multirow{2}{*}{Loc. err.$\downarrow$} & \multirow{2}{*}{Avg. err.$\downarrow$}\\
    ~ & && & Top 3 & & Ratio.$\downarrow$ & & & \\
    \midrule
    &GT &- &$0.002$ &$0.797$ &$9.503$ &-&$0.000$ &$0.000$ &$0.000$\\
    \midrule
    \multirow{8}{*}{\parbox{2.2cm}{\centering \textbf{Train\\On\\Pelvis}}}
    &MDM~\citep{tevet2023human} &\ding{51} &$0.698$ &$0.602$ &$9.197$ &$0.1019$ &$40.22$ &$30.76$ &$59.59$\\
    &PriorMDM~\citep{shafir2024priormdm} &\ding{55} &$0.475$ &$0.583$ &$9.156$ &$0.0897$&$34.57$ &$21.32$ &$44.17$\\
    &GMD~\citep{karunratanakul2023gmd} &\ding{51} &$0.576$ &$0.665$ &$9.206$ &$0.1009$&$9.31$ &$3.21$ &$14.39$\\
    &OmniControl~\citep{xie2024omnicontrol} &\ding{55} &$0.218$ &$0.687$ &$9.422$  &$\mathbf{0.0547}$ &$\underline{3.87}$ &$\underline{0.96}$ &$3.38$\\
    &MotionLCM V2~\citep{dai2024motionlcm} &\ding{55} &$0.531$ &$0.752$ &$9.253$ &$-$ &$18.87$ &$7.69$&$18.97$ \\
    &TLControl~\citep{wan2024tlcontrol}&\ding{55} &$0.271$ &$\underline{0.779}$ &$\underline{9.569}$ &$-$ &$\mathbf{0.00}$ &$\mathbf{0.00}$&$1.08$ \\
    &MaskControl~\citep{pinyoanuntapong2024controlmm}&\ding{55} &$\mathbf{0.061}$ &$\mathbf{0.809}$ &$\mathbf{9.496}$ &$\mathbf{0.0547}$ &$\mathbf{0.00}$ &$\mathbf{0.00}$&$\underline{0.98}$ \\
    &\textbf{ACMDM\text{-}S\text{-}PS22+ProjFlow (ours)} &\ding{51} &$\underline{0.083}$ &$0.755$ &$9.096$ &$\underline{0.0651}$ &$\mathbf{0.00}$ &$\mathbf{0.00}$ &$\mathbf{0.00}$\\
    
    \midrule

    \multirow{5}{*}{\parbox{2.2cm}{\centering \textbf{Train On\\All Joints\\(Average)}}}
    &OmniControl~\citep{xie2024omnicontrol} &\ding{55} &$0.310$ &$0.693$ &$\mathbf{9.502}$ & $\underline{0.0608}$&$\underline{6.17}$ &$\underline{1.07}$ &$4.04$\\
    &TLControl~\citep{wan2024tlcontrol}&\ding{55} &$0.256$ &$\underline{0.782}$ &$9.719$ &$-$ &$\mathbf{0.00}$ &$\mathbf{0.00}$&$1.11$ \\
    &MaskControl~\citep{pinyoanuntapong2024controlmm}&\ding{55} &$\underline{0.083}$ &$\mathbf{0.805}$ &$\underline{9.395}$ &$\mathbf{0.0545}$ &$\mathbf{0.00}$ &$\mathbf{0.00}$&$\underline{0.72}$ \\
    &\textbf{ACMDM\text{-}S\text{-}PS22+ProjFlow (ours)} &\ding{51} &$\mathbf{0.074}$ &$0.752$ &$9.065$ &$0.0624$ &$\mathbf{0.00}$ &$\mathbf{0.00}$ &$\mathbf{0.00}$\\

    \midrule
    \midrule
    & \multirow{2}{*}{Methods} & \multirow{2}{*}{Zero-shot?}& \multirow{2}{*}{FID$\downarrow$}& R-Precision & R-Precision  & R-Precision & \multirow{2}{*}{Matching$\downarrow$} & \multirow{2}{*}{Diversity$\rightarrow$} & \multirow{2}{*}{$-$}\\
    ~ & && & Top 1 &Top 2 & Top 3& & \\
    \midrule
    \multirow{5}{*}{\parbox{2.2cm}{\centering \textbf{UpperBody\\Edit}}} 
    &MDM~\citep{tevet2023human}            &\ding{51}&$4.827$ &$0.298$ &$0.462$ &$0.571$ &$4.598$ &$7.010$& $-$\\
    &OmniControl~\citep{xie2024omnicontrol} &\ding{55}&$1.213$ &$0.374$ &$0.550$ &$0.656$ &$5.228$ &$9.258$& $-$\\
    &MMM~\citep{pinyoanuntapong2024mmm}        &\ding{55}&$0.103$ &$0.500$ &$0.694$ &$\underline{0.798}$ &$2.972$ &$9.254$& $-$\\
    &MotionLCM~\citep{dai2024motionlcm}     &\ding{55}&$0.311$ &$\underline{0.512}$ &$0.685$ &$\underline{0.798}$ &$\underline{2.948}$ &$\underline{9.736}$& $-$\\
    &MaskControl~\citep{pinyoanuntapong2024controlmm}                     &\ding{55}&$\underline{0.074}$ &$\mathbf{0.517}$ &$\mathbf{0.708}$ &$\mathbf{0.804}$ &$\mathbf{2.945}$ &$\mathbf{9.380}$& $-$\\
    &\textbf{ACMDM\text{-}S\text{-}PS22+ProjFlow (ours)} &\ding{51} &$\mathbf{0.051}$ &$0.502$ &$\underline{0.697}$ &$0.793$ &$3.281$ &$10.611$ &$-$\\
    
    \bottomrule
    \end{tabular}}
    \label{tab:control_noAITS263}
\end{table*}

%% file: main.bib
@String(IJCV = {Int. J. Comput. Vis.})

@String(CVPR= {IEEE Conf. Comput. Vis. Pattern Recog.})

@String(ICCV= {Int. Conf. Comput. Vis.})

@String(ECCV= {Eur. Conf. Comput. Vis.})

@String(ICLR = {Int. Conf. Learn. Represent.})

@String(AAAI = {AAAI})

@String(IJCV  = {IJCV})

@String(CVPR  = {CVPR})

@String(ICCV  = {ICCV})

@String(ECCV  = {ECCV})

@String(ICLR  = {ICLR})

@inproceedings{xie2024omnicontrol,
  title     = {OmniControl: Control Any Joint at Any Time for Human Motion Generation},
  author    = {Xie, Yiming and Jampani, Varun and Zhong, Lei and Sun, Deqing and Jiang, Huaizu},
  booktitle = {International Conference on Learning Representations (ICLR)},
  year      = {2024},
  url       = {https://openreview.net/forum?id=gd0lAEtWso}
}

@inproceedings{karunratanakul2023gmd,
  title     = {Guided Motion Diffusion for Controllable Human Motion Synthesis},
  author    = {Karunratanakul, Korrawe and Preechakul, Konpat and Suwajanakorn, Supasorn and Tang, Siyu},
  booktitle = {Proceedings of the IEEE/CVF International Conference on Computer Vision (ICCV)},
  year      = {2023},
  pages     = {21510--21522},
  url       = {https://openaccess.thecvf.com/content/ICCV2023/html/Karunratanakul_Guided_Motion_Diffusion_for_Controllable_Human_Motion_Synthesis_ICCV_2023_paper.html}
}

@article{meng2025acmdm,
  title     = {Absolute Coordinates Make Motion Generation Easy},
  author    = {Meng, Zichong and Han, Zeyu and Peng, Xiaogang and Xie, Yiming and Jiang, Huaizu},
  journal   = {arXiv preprint arXiv:2505.19377},
  year      = {2025},
  url       = {https://arxiv.org/abs/2505.19377}
}

@article{pi2025coda,
  title     = {CoDA: Coordinated Diffusion Noise Optimization for Whole-Body Manipulation of Articulated Objects},
  author    = {Pi, Huaijin and Cen, Zhi and Dou, Zhiyang and Komura, Taku},
  journal   = {arXiv preprint arXiv:2505.21437},
  year      = {2025},
  url       = {https://arxiv.org/abs/2505.21437}
}

@article{ron2025hoidini,
  title     = {HOIDiNi: Human-Object Interaction through Diffusion Noise Optimization},
  author    = {Ron, Roni and Tevet, Guy and Sawdayee, Harel and Bermano, Amit H.},
  journal   = {arXiv preprint arXiv:2506.15625},
  year      = {2025},
  url       = {https://arxiv.org/abs/2506.15625}
}

@inproceedings{dabral2023mofusion,
  title     = {MoFusion: A Framework for Denoising-Diffusion-Based Motion Synthesis},
  author    = {Dabral, Rishabh and Mughal, Muhammad Hamza and Golyanik, Vladislav and Theobalt, Christian},
  booktitle = {CVPR},
  year      = {2023},
  url       = {https://openaccess.thecvf.com/content/CVPR2023/papers/Dabral_MoFusion_A_Framework_for_Denoising-Diffusion-Based_Motion_Synthesis_CVPR_2023_paper.pdf}
}

@article{zhang2024motiondiffuse,
  title     = {Text-Driven Human Motion Generation With Diffusion Model},
  author    = {Zhang, Mingyuan and Cai, Zhongang and Pan, Liang and Hong, Fangzhou and Guo, Xinying and Yang, Lei and Liu, Ziwei},
  journal   = {IEEE TPAMI},
  year      = {2024},
  doi       = {10.1109/TPAMI.2024.3355414},
  url       = {https://doi.org/10.1109/TPAMI.2024.3355414}
}

@inproceedings{shafir2024priormdm,
  title     = {Human Motion Diffusion as a Generative Prior},
  author    = {Shafir, Yoni and Tevet, Guy and Kapon, Roy and Bermano, Amit H.},
  booktitle = {ICLR},
  year      = {2024},
  url       = {https://openreview.net/forum?id=dTpbEdN9kr}
}

@article{dai2024motionlcm,
  title     = {MotionLCM: Real-time Controllable Motion Generation via Latent Consistency Models},
  author    = {Dai, Wenxun and others},
  journal   = {arXiv preprint arXiv:2404.19759},
  year      = {2024},
  url       = {https://arxiv.org/abs/2404.19759}
}

@inproceedings{flow_lipman2023,
  author    = {Yaron Lipman and Ricky T. Q. Chen and Heli Ben-Hamu and Maximilian Nickel and Matthew Le},
  title     = {Flow Matching for Generative Modeling},
  booktitle = {International Conference on Learning Representations (ICLR)},
  year      = {2023},
  url       = {https://openreview.net/forum?id=PqvMRDCJT9t},
  eprint    = {2210.02747},
  archivePrefix = {arXiv}
}

@inproceedings{flow_liu2023,
  author    = {Xingchao Liu and Chengyue Gong and Qiang Liu},
  title     = {Flow Straight and Fast: Learning to Generate and Transfer Data with Rectified Flow},
  booktitle = {International Conference on Learning Representations (ICLR)},
  year      = {2023},
  url       = {https://openreview.net/forum?id=XVjTT1nw5z},
  eprint    = {2209.03003},
  archivePrefix = {arXiv}
}

@inproceedings{flow_albergo2023,
  author    = {Michael S. Albergo and Eric Vanden{-}Eijnden},
  title     = {Building Normalizing Flows with Stochastic Interpolants},
  booktitle = {International Conference on Learning Representations (ICLR)},
  year      = {2023},
  url       = {https://openreview.net/forum?id=li7qeBbCR1t},
  eprint    = {2209.15571},
  archivePrefix = {arXiv}
}

@inproceedings{ddpm_ho2020,
  author    = {Jonathan Ho and Ajay Jain and Pieter Abbeel},
  title     = {Denoising Diffusion Probabilistic Models},
  booktitle = {Advances in Neural Information Processing Systems},
  year      = {2020},
  volume    = {33},
  pages     = {6840--6851},
  publisher = {Curran Associates, Inc.},
  editor    = {Hugo Larochelle and Marc'Aurelio Ranzato and Raia Hadsell and Maria{-}Florina Balcan and Hsuan{-}Tien Lin},
  url       = {https://proceedings.neurips.cc/paper/2020/hash/4c5bcfec8584af0d967f1ab10179ca4b-Abstract.html},
  eprint    = {2006.11239},
  archivePrefix = {arXiv}
}

@inproceedings{score_song2021,
  author    = {Yang Song and Jascha Sohl{-}Dickstein and Diederik P. Kingma and Abhishek Kumar and Stefano Ermon and Ben Poole},
  title     = {Score{-}Based Generative Modeling through Stochastic Differential Equations},
  booktitle = {International Conference on Learning Representations (ICLR)},
  year      = {2021},
  url       = {https://openreview.net/forum?id=PxTIG12RRHS},
  eprint    = {2011.13456},
  archivePrefix = {arXiv}
}

@inproceedings{sd3_esser2024,
  author    = {Patrick Esser and Sumith Kulal and Andreas Blattmann and Rahim Entezari and Jonas M{\"u}ller and Harry Saini and Yam Levi and Dominik Lorenz and Axel Sauer and Frederic Boesel and Dustin Podell and Tim Dockhorn and Zion English and Robin Rombach},
  title     = {Scaling Rectified Flow Transformers for High-Resolution Image Synthesis},
  booktitle = {Proceedings of the 41st International Conference on Machine Learning (ICML)},
  year      = {2024},
  series    = {Proceedings of Machine Learning Research},
  volume    = {235},
  pages     = {12606--12633},
  address   = {Vienna, Austria},
  month     = {21--27 Jul},
  publisher = {PMLR},
  url       = {https://proceedings.mlr.press/v235/esser24a.html}
}

@misc{flow_lipman2024,
  author       = {Yaron Lipman and Marton Havasi and Peter Holderrieth and Neta Shaul and Matt Le and Brian Karrer and Ricky T. Q. Chen and David Lopez{-}Paz and Heli Ben{-}Hamu and Itai Gat},
  title        = {Flow Matching Guide and Code},
  year         = {2024},
  eprint       = {2412.06264},
  archivePrefix= {arXiv},
  primaryClass = {cs.LG},
  doi          = {10.48550/arXiv.2412.06264},
  url          = {https://arxiv.org/abs/2412.06264}
}

@misc{kim2025flowdps,
  title         = {{FlowDPS}: Flow-Driven Posterior Sampling for Inverse Problems},
  author        = {Kim, Jeongsol and Kim, Bryan Sangwoo and Ye, Jong Chul},
  year          = {2025},
  eprint        = {2503.08136},
  archivePrefix = {arXiv},
  primaryClass  = {cs.CV},
  doi           = {10.48550/arXiv.2503.08136},
  url           = {https://arxiv.org/abs/2503.08136}
}

@inproceedings{tevet2023human,
  title     = {Human Motion Diffusion Model},
  author    = {Tevet, Guy and Raab, Sigal and Gordon, Brian and Shafir, Yoni and Cohen-Or, Daniel and Bermano, Amit Haim},
  booktitle = {International Conference on Learning Representations (ICLR)},
  year      = {2023}
}

@inproceedings{zhang2023controlnet,
  title     = {Adding Conditional Control to Text-to-Image Diffusion Models},
  author    = {Zhang, Lvmin and Rao, Anyi and Agrawala, Maneesh},
  booktitle = {Proceedings of the IEEE/CVF International Conference on Computer Vision (ICCV)},
  pages     = {3813--3824},
  year      = {2023}
}

@InProceedings{karunratanakul2024dno,
    author    = {Karunratanakul, Korrawe and Preechakul, Konpat and Aksan, Emre and Beeler, Thabo and Suwajanakorn, Supasorn and Tang, Siyu},
    title     = {Optimizing Diffusion Noise Can Serve As Universal Motion Priors},
    booktitle = {Proceedings of the IEEE/CVF Conference on Computer Vision and Pattern Recognition (CVPR)},
    month     = {June},
    year      = {2024},
    pages     = {1334-1345}
}

@Article{zhong2025sketch2anim,
  Title     = {Sketch2Anim: Towards Transferring Sketch Storyboards into 3D Animation},
  Author    = {Zhong, Lei and Guo, Chuan and Xie, Yiming and Wang, Jiawei and Li, Changjian},
  Journal   = {ACM Transactions on Graphics},
  volume    = {44},
  number    = {4},
  pages     = {1--15},
  Year      = {2025},
  Publisher = {ACM},
  address   = {New York, NY, USA}
}

@article{wang2022ddnm,
  title={Zero-Shot Image Restoration Using Denoising Diffusion Null-Space Model},
  author={Wang, Yinhuai and Yu, Jiwen and Zhang, Jian},
  journal={The Eleventh International Conference on Learning Representations},
  year={2023}
}

@InProceedings{guo2022humanml3d,
    author    = {Guo, Chuan and Zou, Shihao and Zuo, Xinxin and Wang, Sen and Ji, Wei and Li, Xingyu and Cheng, Li},
    title     = {Generating Diverse and Natural 3D Human Motions From Text},
    booktitle = {Proceedings of the IEEE/CVF Conference on Computer Vision and Pattern Recognition (CVPR)},
    month     = {June},
    year      = {2022},
    pages     = {5152-5161}
}

@conference{mahmood2019amass,
  title = {{AMASS}: Archive of Motion Capture as Surface Shapes},
  author = {Mahmood, Naureen and Ghorbani, Nima and Troje, Nikolaus F. and Pons-Moll, Gerard and Black, Michael J.},
  booktitle = {International Conference on Computer Vision},
  pages = {5442--5451},
  month = oct,
  year = {2019},
  month_numeric = {10}
}

@inproceedings{guo2020action2motion,
  title={Action2motion: Conditioned generation of 3d human motions},
  author={Guo, Chuan and Zuo, Xinxin and Wang, Sen and Zou, Shihao and Sun, Qingyao and Deng, Annan and Gong, Minglun and Cheng, Li},
  booktitle={Proceedings of the 28th ACM International Conference on Multimedia},
  pages={2021--2029},
  year={2020}
}

@inproceedings{studer2024factorized,
author = {Studer, Justin and Agrawal, Dhruv and Borer, Dominik and Sadat, Seyedmorteza and Sumner, Robert W. and Guay, Martin and Buhmann, Jakob},
title = {Factorized Motion Diffusion for Precise and Character-Agnostic Motion Inbetweening},
year = {2024},
isbn = {9798400710902},
publisher = {Association for Computing Machinery},
address = {New York, NY, USA},
url = {https://doi.org/10.1145/3677388.3696338},
doi = {10.1145/3677388.3696338},
booktitle = {Proceedings of the 17th ACM SIGGRAPH Conference on Motion, Interaction, and Games},
articleno = {11},
numpages = {10},
location = {Arlington, VA, USA},
series = {MIG '24}
}

@article{agrawal2024skelbetweener,
author = {Agrawal, Dhruv and Buhmann, Jakob and Borer, Dominik and Sumner, Robert W. and Guay, Martin},
title = {SKEL-Betweener: a Neural Motion Rig for Interactive Motion Authoring},
year = {2024},
issue_date = {December 2024},
publisher = {Association for Computing Machinery},
address = {New York, NY, USA},
volume = {43},
number = {6},
issn = {0730-0301},
url = {https://doi.org/10.1145/3687941},
doi = {10.1145/3687941},
journal = {ACM Trans. Graph.},
month = nov,
articleno = {247},
numpages = {11}
}

@InProceedings{meng2025rethinking,
    author    = {Meng, Zichong and Xie, Yiming and Peng, Xiaogang and Han, Zeyu and Jiang, Huaizu},
    title     = {Rethinking Diffusion for Text-Driven Human Motion Generation: Redundant Representations, Evaluation, and Masked Autoregression},
    booktitle = {Proceedings of the IEEE/CVF Conference on Computer Vision and Pattern Recognition (CVPR)},
    month     = {June},
    year      = {2025},
    pages     = {27859-27871}
}

@article{hu2023mfm,
  title   = {Motion Flow Matching for Human Motion Synthesis and Editing},
  author  = {Hu, Vincent Tao and Yin, Wenzhe and Ma, Pingchuan and Chen, Yunlu and Fernando, Basura and Asano, Yuki M. and Gavves, Efstratios and Mettes, Pascal and Ommer, Bj{\"o}rn and Snoek, Cees G. M.},
  journal = {arXiv preprint arXiv:2312.08895},
  year    = {2023},
  url     = {https://arxiv.org/abs/2312.08895}
}

@article{cuba2025flowmotion,
  title   = {FlowMotion: Target-Predictive Conditional Flow Matching for Jitter-Reduced Text-Driven Human Motion Generation},
  author  = {Canales Cuba, Manolo and do Carmo Mel{\'{\i}}cio, Vin{\'{\i}}cius and Gois, Jo{\~a}o Paulo},
  journal = {arXiv preprint arXiv:2504.01338},
  year    = {2025},
  url     = {https://arxiv.org/abs/2504.01338}
}

@inproceedings{wan2024tlcontrol,
  title     = {TLControl: Trajectory and Language Control for Human Motion Synthesis},
  author    = {Wan, Weilin and Dou, Zehao and Komura, Taku and Wang, Wenping and Jayaraman, Dinesh and Liu, Lingjie},
  booktitle = {ECCV},
  year      = {2024},
  doi       = {10.1007/978-3-031-72913-3_3},
  url       = {https://arxiv.org/abs/2311.17135}
}

@inproceedings{liang2024omg,
  title     = {OMG: Towards Open-vocabulary Motion Generation via Mixture of Controllers},
  author    = {Liang, Han and Bao, Jiacheng and Zhang, Ruichi and Ren, Sihan and Xu, Yuecheng and Yang, Sibei and Chen, Xin and Yu, Jingyi and Xu, Lan},
  booktitle = {CVPR},
  year      = {2024},
  url       = {https://openaccess.thecvf.com/content/CVPR2024/papers/Liang_OMG_Towards_Open-vocabulary_Motion_Generation_via_Mixture_of_Controllers_CVPR_2024_paper.pdf}
}

@inproceedings{zhang2023remodiffuse,
  title     = {ReMoDiffuse: Retrieval-Augmented Motion Diffusion Model},
  author    = {Zhang, Mingyuan and Guo, Xinying and Pan, Liang and Cai, Zhongang and Hong, Fangzhou and Li, Huirong and Yang, Lei and Liu, Ziwei},
  booktitle = {ICCV},
  year      = {2023},
  pages     = {364--373},
  url       = {https://openaccess.thecvf.com/content/ICCV2023/html/Zhang_ReMoDiffuse_Retrieval-Augmented_Motion_Diffusion_Model_ICCV_2023_paper.html}
}

@inproceedings{rempe2023tracepace,
  title     = {Trace and Pace: Controllable Pedestrian Animation via Guided Trajectory Diffusion},
  author    = {Rempe, Davis and Luo, Zhengyi and Peng, Xue Bin and Yuan, Ye and Kitani, Kris and Kreis, Karsten and Fidler, Sanja and Litany, Or},
  booktitle = {CVPR},
  year      = {2023},
  url       = {https://openaccess.thecvf.com/content/CVPR2023/papers/Rempe_Trace_and_Pace_Controllable_Pedestrian_Animation_via_Guided_Trajectory_Diffusion_CVPR_2023_paper.pdf}
}

@article{pinyoanuntapong2024controlmm,
  title   = {ControlMM: Controllable Masked Motion Generation},
  author  = {Pinyoanuntapong, Ekkasit and Saleem, Muhammad Usama and Karunratanakul, Korrawe and Wang, Pu and Xue, Hongfei and Chen, Chen and Guo, Chuan and Cao, Junli and Ren, Jian and Tulyakov, Sergey},
  journal = {arXiv preprint arXiv:2410.10780},
  year    = {2024},
  url     = {https://arxiv.org/abs/2410.10780}
}

@inproceedings{serifi2024robotmdm,
  title     = {Robot Motion Diffusion Model: Motion Generation for Robotic Characters},
  author    = {Serifi, Agon and Grandia, Ruben and Knoop, Espen and Gross, Markus and B{\"a}cher, Moritz},
  booktitle = {SIGGRAPH Asia 2024 Conference Papers},
  year      = {2024},
  publisher = {ACM},
  doi       = {10.1145/3680528.3687626}
}

@inproceedings{alegre2025amor,
  title     = {AMOR: Adaptive Character Control through Multi-Objective Reinforcement Learning},
  author    = {Alegre, Lucas N. and Serifi, Agon and Grandia, Ruben and M{\"u}ller, David and Knoop, Espen and B{\"a}cher, Moritz},
  booktitle = {SIGGRAPH 2025 Conference Papers},
  year      = {2025},
  publisher = {ACM},
  doi       = {10.1145/3721238.3730656}
}

@inproceedings{lee2019dancing,
  title     = {Dancing to Music},
  author    = {Lee, Hsin-Ying and Yang, Xiaodong and Liu, Ming-Yu and Wang, Ting-Chun and Lu, Yu-Ding and Yang, Ming-Hsuan and Kautz, Jan},
  booktitle = {NeurIPS},
  year      = {2019}
}

@inproceedings{li2021danceformer,
  title     = {DanceFormer: Music Conditioned 3D Dance Generation with Parametric Motion Transformer},
  author    = {Li, Buyu and Zhao, Yongchi and Shi, Zhelun and Sheng, Lu},
  booktitle = {AAAI},
  year      = {2021}
}

@inproceedings{li2021aichoreographer,
  title     = {AI Choreographer: Music Conditioned 3D Dance Generation with AIST++},
  author    = {Li, Ruilong and Yang, Sha and Ross, David A. and Kanazawa, Angjoo},
  booktitle = {ICCV},
  year      = {2021}
}

@inproceedings{li2022bailando,
  title     = {Bailando: 3D Dance Generation by Actor-Critic GPT with Choreographic Memory},
  author    = {Li, Siyao and Yu, Weijiang and Gu, Tianpei and Lin, Chunze and Wang, Quan and Qian, Chen and Loy, Chen Change and Liu, Ziwei},
  booktitle = {CVPR},
  year      = {2022}
}

@article{li2023bailandopp,
  title   = {Bailando++: 3D Dance GPT with Choreographic Memory},
  author  = {Li, Siyao and Yu, Weijiang and Gu, Tianpei and Lin, Chunze and Wang, Quan and Qian, Chen and Loy, Chen Change and Liu, Ziwei},
  journal = {IEEE TPAMI},
  year    = {2023}
}

@inproceedings{tseng2022edge,
  title     = {EDGE: Editable Dance Generation From Music},
  author    = {Tseng, Jo-Han and Castellon, Rodrigo and Liu, C. Karen},
  booktitle = {CVPR},
  year      = {2022}
}

@inproceedings{cha2024text2hoi,
  title     = {Text2HOI: Text-guided 3D Motion Generation for Hand-Object Interaction},
  author    = {Cha, Junuk and Kim, Jihyeon and Yoon, Jae Shin and Baek, Seungryul},
  booktitle = {CVPR},
  year      = {2024}
}

@inproceedings{diller2024cghoi,
  title     = {CG-HOI: Contact-Guided 3D Human-Object Interaction Generation},
  author    = {Diller, Christian and Dai, Angela},
  booktitle = {CVPR},
  year      = {2024}
}

@inproceedings{kulkarni2024nifty,
  title     = {NIFTY: Neural Object Interaction Fields for Guided Human Motion Synthesis},
  author    = {Kulkarni, Nilesh and Rempe, Davis and Genova, Kyle and Kundu, Abhijit and Johnson, Justin and Fouhey, David and Guibas, Leonidas},
  booktitle = {CVPR},
  year      = {2024}
}

@article{li2023controllableHOI,
  title   = {Controllable Human-Object Interaction Synthesis},
  author  = {Li, Jiaman and Clegg, Alexander and Mottaghi, Roozbeh and Wu, Jiajun and Puig, Xavier and Liu, C. Karen},
  journal = {arXiv:2312.03913},
  year    = {2023}
}

@inproceedings{du2023avatars,
  title     = {Avatars Grow Legs: Generating Smooth Human Motion from Sparse Tracking Inputs with Diffusion Models},
  author    = {Du, Yuming and Kips, Robin and Pumarola, Albert and Starke, Sebastian and Thabet, Ali and Sanakoyeu, Artsiom},
  booktitle = {CVPR},
  year      = {2023}
}

@inproceedings{huang2023diffusion3d,
  title     = {Diffusion-based Generation, Optimization, and Planning in 3D Scenes},
  author    = {Huang, Siyuan and Wang, Zan and Li, Puhao and Jia, Baoxiong and Liu, Tengyu and Zhu, Yixin and Liang, Wei and Zhu, Song-Chun},
  booktitle = {CVPR},
  year      = {2023}
}

@inproceedings{wang2024moveasyousay,
  title     = {Move as You Say, Interact as You Can: Language-Guided Human Motion Generation with Scene Affordance},
  author    = {Wang, Zan and Chen, Yixin and Jia, Baoxiong and Li, Puhao and Zhang, Jinlu and Zhang, Jingze and Liu, Tengyu and Zhu, Yixin and Liang, Wei and Huang, Siyuan},
  booktitle = {CVPR},
  year      = {2024}
}

@inproceedings{liu2024progmogen,
  title     = {Programmable Motion Generation for Open-set Motion Control Tasks},
  author    = {Liu, Hanchao and Zhan, Xiaohang and Huang, Shaoli and Mu, Tai-Jiang and Shan, Ying},
  booktitle = {CVPR},
  year      = {2024}
}

@article{zhong2024smoodi,
  title   = {SMOODI: Stylized Motion Diffusion Model},
  author  = {Zhong, Lei and Xie, Yiming and Jampani, Varun and Sun, Deqing and Jiang, Huaizu},
  journal = {arXiv:2407.12783},
  year    = {2024}
}

@inproceedings{diomataris2024wandr,
  title     = {WANDR: Intention-Guided Human Motion Generation},
  author    = {Diomataris, Markos and Athanasiou, Nikos and Taheri, Omid and Wang, Xi and Hilliges, Otmar and Black, Michael J.},
  booktitle = {CVPR},
  year      = {2024}
}

@inproceedings{petrovich2024multitrack,
    author    = {Petrovich, Mathis and Litany, Or and Iqbal, Umar and Black, Michael J. and Varol, Gul and Bin Peng, Xue and Rempe, Davis},
    title     = {Multi-Track Timeline Control for Text-Driven 3D Human Motion Generation},
    booktitle = {Proceedings of the IEEE/CVF Conference on Computer Vision and Pattern Recognition (CVPR) Workshops},
    month     = {June},
    year      = {2024},
    pages     = {1911-1921}
}

@inproceedings{cohan2024condmdi,
  title     = {Flexible Motion In-betweening with Diffusion Models},
  author    = {Cohan, Setareh and Reda, Daniele and Tevet, Guy and Peng, Xue Bin and van de Panne, Michiel},
  booktitle = {SIGGRAPH 2024 Conference Papers},
  year      = {2024},
  doi       = {10.1145/3641519.3657414},
  url       = {https://xbpeng.github.io/projects/CondMDI/CondMDI_2024.pdf},
  publisher = {ACM}
}

@inproceedings{
wang2023zeroshot,
title={Zero-Shot Image Restoration Using Denoising Diffusion Null-Space Model},
author={Yinhuai Wang and Jiwen Yu and Jian Zhang},
booktitle={The Eleventh International Conference on Learning Representations },
year={2023},
url={https://openreview.net/forum?id=mRieQgMtNTQ}
}

@inproceedings{
song2023pseudoinverseguided,
title={Pseudoinverse-Guided Diffusion Models for Inverse Problems},
author={Jiaming Song and Arash Vahdat and Morteza Mardani and Jan Kautz},
booktitle={International Conference on Learning Representations},
year={2023},
url={https://openreview.net/forum?id=9_gsMA8MRKQ}
}

@inproceedings{rout2023solving,
title={Solving Linear Inverse Problems Provably via Posterior Sampling with Latent Diffusion Models},
author={Litu Rout and Negin Raoof and Giannis Daras and Constantine Caramanis and Alex Dimakis and Sanjay Shakkottai},
booktitle={Thirty-seventh Conference on Neural Information Processing Systems},
year={2023},
url={https://openreview.net/forum?id=XKBFdYwfRo}
}

@inproceedings{
    song2024solving,
    title={Solving Inverse Problems with Latent Diffusion Models via Hard Data Consistency},
    author={Bowen Song and Soo Min Kwon and Zecheng Zhang and Xinyu Hu and Qing Qu and Liyue Shen},
    booktitle={The Twelfth International Conference on Learning Representations},
    year={2024},
    url={https://openreview.net/forum?id=j8hdRqOUhN}
}

@article{zhang2024improving,
  title={Improving diffusion inverse problem solving with decoupled noise annealing},
  author={Zhang, Bingliang and Chu, Wenda and Berner, Julius and Meng, Chenlin and Anandkumar, Anima and Song, Yang},
  journal={arXiv preprint arXiv:2407.01521},
  year={2024}
}

@inproceedings{
kawar2022denoising,
title={Denoising Diffusion Restoration Models},
author={Bahjat Kawar and Michael Elad and Stefano Ermon and Jiaming Song},
booktitle={Advances in Neural Information Processing Systems},
editor={Alice H. Oh and Alekh Agarwal and Danielle Belgrave and Kyunghyun Cho},
year={2022},
url={https://openreview.net/forum?id=kxXvopt9pWK}
}

@inproceedings{
chung2023diffusion,
title={Diffusion Posterior Sampling for General Noisy Inverse Problems},
author={Hyungjin Chung and Jeongsol Kim and Michael Thompson Mccann and Marc Louis Klasky and Jong Chul Ye},
booktitle={International Conference on Learning Representations},
year={2023},
url={https://openreview.net/forum?id=OnD9zGAGT0k}
}

@inproceedings{patel2024steering,
    author    = {Patel, Maitreya and Wen, Song and Metaxas, Dimitris N. and Yang, Yezhou},
    title     = {FlowChef: Steering of Rectified Flow Models for Controlled Generations},
    booktitle = {Proceedings of the IEEE/CVF International Conference on Computer Vision (ICCV)},
    month     = {October},
    year      = {2025},
    pages     = {15308-15318}
}

@inproceedings{
martin2025pnpflow,
title={PnP-Flow: Plug-and-Play Image Restoration with Flow Matching},
author={S{\'e}gol{\`e}ne Tiffany Martin and Anne Gagneux and Paul Hagemann and Gabriele Steidl},
booktitle={The Thirteenth International Conference on Learning Representations},
year={2025},
url={https://openreview.net/forum?id=5AtHrq3B5R}
}

@inproceedings{
tanaka2023interaction,
title={Role-aware Interaction Generation from Textual Description},
author={Mikihiro Tanaka and Kent Fujiwara},
booktitle=ICCV,
year={2023}
}

@article{liang2024intergen,
  author    = {Han Liang and Wenqian Zhang and Wenxuan Li and Jingyi Yu and Lan Xu},
  title     = {Intergen: Diffusion-based Multi-human Motion Generation Under Complex Interactions},
  journal   = IJCV,
  pages     = {1--21},
  year      = {2024},
  publisher = {Springer}
}

@inproceedings{fan2024freemotion,
author    = {Ke Fan and Junshu Tang and Weijian Cao and Ran Yi and Moran Li and Jingyu Gong and Jiangning Zhang and Yabiao Wang and Chengjie Wang and Lizhuang Ma},
title     = {FreeMotion: A Unified Framework for Number-free Text-to-Motion Synthesis},
year={2024},
booktitle=ECCV,
}

@inproceedings{ota2025pino,
  title={Pino: Person-interaction noise optimization for long-duration and customizable motion generation of arbitrary-sized groups},
  author={Ota, Sakuya and Yu, Qing and Fujiwara, Kent and Ikehata, Satoshi and Sato, Ikuro},
  booktitle={ICCV},
  year={2025}
}

@InProceedings{pinyoanuntapong2024mmm,
    author    = {Pinyoanuntapong, Ekkasit and Wang, Pu and Lee, Minwoo and Chen, Chen},
    title     = {MMM: Generative Masked Motion Model},
    booktitle = {Proceedings of the IEEE/CVF Conference on Computer Vision and Pattern Recognition (CVPR)},
    month     = {June},
    year      = {2024},
    pages     = {1546-1555}
}
